\documentclass[runningheads]{llncs}
\usepackage{graphicx}
\usepackage{mathtools}
\usepackage{amsmath,amssymb,amsfonts}
\usepackage{stmaryrd}

\usepackage{graphicx}
\usepackage{textcomp}
\usepackage{xcolor}
\graphicspath{ {./images/} }
\usepackage{tikz}
\usepackage{siunitx,booktabs}
\usepackage{multirow}
\usepackage{amsmath}
\usepackage{algorithm}
\usepackage[noend]{algpseudocode}
\usepackage{hyperref}
\usepackage{subfig}

\usetikzlibrary{calc}

\usepackage{array}
\newcolumntype{P}[1]{>{\centering\arraybackslash}p{#1}}
  
\begin{document}
\title{Smart-Tree: Neural Medial Axis Approximation of Point Clouds for 3D Tree Skeletonization \thanks{Supported by University of Canterbury, New Zealand.}}
\titlerunning{Smart-Tree: Neural Axis Approximation}
\author{Harry Dobbs\inst{1}\orcidID{0000-0003-3685-940X}\and
Oliver Batchelor\inst{2}\orcidID{0000-0002-6542-1661}\and
Richard Green\inst{2}\orcidID{0000-0001-5149-722X} \and
James Atlas\inst{2}\orcidID{0000-0002-8030-6098}}

\authorrunning{H. Dobbs et al.}
\institute{UC Vision Research Lab, \\ University of Canterbury, \\ Christchurch 8041, New Zealand \\
\email{harry.dobbs@pg.canterbury.ac.nz} \\
\email{\{oliver.batchelor, richard.green, james.atlas\}@canterbury.ac.nz} \\ 
\url{https://ucvision.org.nz/}}
%\and
%Springer Heidelberg, Tiergartenstr. 17, 69121 Heidelberg, Germany
\maketitle              % typeset the header of the contribution
\begin{abstract}
This paper introduces Smart-Tree, a supervised method for approximating the medial axes of branch skeletons from a tree point cloud. Smart-Tree uses a sparse voxel convolutional neural network to extract the radius and direction towards the medial axis of each input point. A greedy algorithm performs robust skeletonization using the estimated medial axis. Our proposed method provides robustness to complex tree structures and improves fidelity when dealing with self-occlusions, complex geometry, touching branches, and varying point densities. We evaluate Smart-Tree using a multi-species synthetic tree dataset and perform qualitative analysis on a real-world tree point cloud. Our experimentation with synthetic and real-world datasets demonstrates the robustness of our approach over the current state-of-the-art method. The dataset\footnote{\url{https://github.com/uc-vision/synthetic-trees}} and source code\footnote{\url{https://github.com/uc-vision/smart-tree}} are publicly available. 

\end{abstract}

\section{Introduction} \label{intro_section}

Digital tree models have many applications, such as biomass estimation \cite{fan2020adqsm,kankare2013individual,fan2020new}, growth modelling \cite{tompalski2021estimating,spalding2013image,chaudhury2018machine}, forestry management \cite{white2013utility,molina2022operationalizing,calders2020terrestrial}, urban microclimate simulation \cite{xu20213d}, and agri-tech applications, such as robotic pruning \cite{zahid2021technological,botterill2017robot}, and fruit picking \cite{arikapudi2021robotic}. 

A comprehensive digital tree model relies on the ability to extract a skeleton from a point cloud. In general, a skeleton is a thin structure that encodes an object's topology and basic geometry \cite{chaudhury2020skeletonization}.
Skeleton extraction from 3D point clouds has been studied extensively in computer vision and graphics literature \cite{saha2016survey} for understanding shapes and topology. Applied to tree point clouds, this is a challenging problem due to self-occlusions, complex geometry, touching branches, and varying point densities.

There are many existing approaches for recovering tree skeletons from point clouds (recent survey \cite{cardenas2022modeling}). These approaches can be categorized as follows; neighbourhood graph, medial axis approximation, voxel and mathematical morphology, and segmentation. 

Neighbourhood graph methods use K-nearest neighbours (usually within a search radius) or Delaunay triangulation to create an initial graph from the point cloud. Multiple implementations \cite{verroust1999extracting,xu2007knowledge,delagrange2014pypetree,hackenberg2015simpletree}, then use this graph to quantize the points into bins based on the distance from the root node. The bin centroids are connected based on constraints to create a skeleton.  
Livny et al. \cite{livny2010automatic} use the neighbourhood graph to perform global optimizations based on a smoothed orientation field. 
Du et al. \cite{du2019adtree} use the graph to construct a Minimum Spanning Tree (MST) and perform iterative graph simplification to extract a skeleton. However, these methods have some drawbacks. Gaps in the point cloud, such as those caused by occlusions, can lead to a disconnected neighbourhood graph. Additionally, false connections may occur, especially when branches are close to one another.

Medial axis approximation works by estimating the medial surface and then thinning it to a medial axis. An approach in the \emph{L1-Medial Skeleton} method \cite{huang2013l1} was proposed for point clouds by iterative sampling and redistributing points to the sample centre. Similarly, Cao et al. \cite{cao2010point} proposed using Laplacian-based contraction to estimate the medial axis. However, these methods are sensitive to irregularities in the point cloud and require a densely sampled object as input.

Voxel and mathematical morphology were implemented in \cite{gorte2004structuring} and later refined in \cite{gorte2006skeletonization}. The point cloud is converted into a 3D voxel grid and then undergoes opening and closing, resulting in a thinned voxel model.
The voxel spatial resolution is a key parameter, as a too-fine resolution will lead to many holes. In contrast, a larger resolution can lose significant detail as multiple points become a single voxel (different branches may all become connected). A further limitation of this method is the required memory and time, which increases with the third power of the resolution. This method only aims to find prominent structures of trees rather than finer branches.

Segmentation approaches work by segmenting points into groups from the same branch. Raumonen \cite{raumonen2013fast} et al. create surface patches along the tree and then grow these patches into branches. However, this method assumes that local areas of the tree have a uniform density.

Recently a deep learning segmentation approach was proposed in TreePartNet \cite{liu2021treepartnet}. This method uses two networks, one to detect branching points and another to detect cylindrical representations. It requires a sufficiently sampled point cloud as it relies on the ability to detect junctions accurately and embed local point groups. However, it struggles to work on larger point clouds due to the memory constraints of PointNet++ \cite{qi2017pointnet++}. Numerous network architectures can process point clouds \cite{guo2020deep}; however, point clouds in general, but in particular, trees are spatially large and sparse, containing fine details - for this reason, we utilise submanifold sparse CNNs \cite{graham2017submanifold,spconv2022,tang2022torchsparse,choy20194d}. 

We propose a deep-learning-based method to estimate the medial axis of a tree point cloud. A neighbourhood graph approach is then used to extract the skeleton. This method mitigates the effects of errors commonly caused when neighbouring branches get close or overlap, as well as improving robustness to common challenges such as varying point density, noise and missing data. Our contributions are as follows:
\begin{itemize}
    \item A synthetic point cloud generation tool was developed for creating a wide range of labelled tree point clouds.
    \item A demonstration of how a sparse convolutional neural network can effectively predict the position of the medial axis.
    \item A skeletonization algorithm is implemented that can use the information from the neural network to perform a robust skeletonization.
    \item The method is evaluated against the state-of-the-art automatic skeletonization method.
    \item The method's ability to generalize is demonstrated by applying it to real data.
\end{itemize}

\section{Method} \label{method_section}

\subsection{Dataset}
We use synthetic data for multiple reasons. First, it has a known ground truth skeleton for quantitative evaluation. Second, we can efficiently label the point clouds - allowing us to generate data for a broad range of species.

We create synthetic trees using a tree modelling software called SpeedTree \cite{speedtree}. For evaluation, we generate the tree meshes without foliage, which otherwise increase the level of occlusion. In the general case, we remove foliage using a segmentation step. Generating point clouds from the tree meshes is done by emulating a spiral drone flight path around each tree and capturing RGBD images at a resolution of 2.1 megapixels (1920 x 1080 pixels). 

We randomly select a sky-box for each tree for varying lighting conditions. The depth maps undergo augmentations such as jitter, dilation and erosion to replicate photogrammetry noise. We extract a point cloud from the fused RGBD images. We remove duplicate points by performing a voxel downsample at a 1cm resolution. The final point clouds have artefacts such as varying point density, missing areas (some due to self-occlusion) and noise.  

We select six species with SpeedTree models. We produce 100 point clouds (600 total) for each of the six tree species, which vary in intricacy and size. Of these, 480 are used for training, while 60 are reserved for validation and 60 for testing. We show images of the synthetic point cloud dataset in the results section (Section \ref{results_section}). Future revisions will include point clouds with foliage and a wider variety of species.

\subsection{Skeletonization Overview}
Our skeletonization method comprises several stages shown in Figure \ref{fig:pipeline}. We use labelled point clouds to train a sparse convolutional neural network to predict each input point’s radius and direction toward the medial axis (ground truth skeleton). Using these predictions, we then translate surface points to estimated medial axis positions and construct a constrained neighbourhood graph. We extract the skeleton using a greedy algorithm to find paths from the root to terminal points. The neural network predictions help to avoid ambiguities with unknown branch radii and separate points which would be close in proximity but from different branches.

% \begin{figure}
% \centering
% \begin{tabular}{cccccc}
%   \includegraphics[height=22mm, trim={10cm 0 10cm 0},clip]{images/species-point-clouds/cherry.png} &  
%   \includegraphics[height=22mm, trim={10cm 0 10cm 0},clip]{images/species-point-clouds/walnut.png} &
%   \includegraphics[height=22mm, trim={10cm 0 10cm 0},clip]{images/species-point-clouds/pine.png} \\
%   \includegraphics[height=22mm, trim={10cm 0 10cm 0},clip]{images/species-point-clouds/eucalyptus.png} &
%   \includegraphics[height=22mm, trim={10cm 0 10cm 0},clip]{images/species-point-clouds/apple.png} &
%   \includegraphics[height=22mm, trim={10cm 0 10cm 0},clip]{images/species-point-clouds/ginkgo.png} \\
% \end{tabular}
% \caption{Synthetic Point Clouds (clockwise): Cherry (a), Walnut (b), Pine (c), Eucalyptus (d), Apple (e), Ginkgo (f).}
% \label{fig:synthetic-pcds}
% \vspace*{-0.5cm}
% \end{figure}

%A summary of the dataset is shown in Table \ref{dataset-statstics}
% \begin{table}[h]
% \centering
% \caption{Dataset Statistics}
% \label{dataset-statstics}
% %\vspace{0.5cm}
% \begin{tabular*}{\textwidth}{    
%     P{0.25\textwidth}  |               
%     P{0.25\textwidth}  |        
%     P{0.25\textwidth}  |             
%     P{0.25\textwidth}                
% }
% Species & Height & Number of Points & Complexity \\
% \hline
% Cherry & 0.95 & 0.86 & 0.90 \\
% Eucalyptus & 0.98 & 0.99  & 0.98 \\
% Apple & 0.98 & 0.99  & 0.98 \\
% Walnut & 0.98 & 0.99  & 0.98 \\
% Pine & 0.98 & 0.99  & 0.98 \\
% Ginkgo & 0.98 & 0.99  & 0.98 \\
% \end{tabular*}
% \end{table}

\subsection{Neural Network}
Our network takes an input set of $N$ arbitrary points $\left\{Pi | i = 1, ..., N \right\}$, where each point $Pi$ is a vector of its $(x, y, z)$ coordinates plus additional features such as colour $(r, g,b)$. Each point is voxelized at a resolution of 1cm. Our proposed network will then, for each voxelized point, learn an associated radius $\left\{Ri | i = 1, ..., N \right\}$ where $Ri$ is a vector of corresponding radii and a direction vector $\left\{Di | i = 1, ..., N \right\}$ where $Di$ is a normalized direction vector pointing towards the medial axis. 

The network is implemented as a submanifold sparse CNN using SpConv \cite{spconv2022}, and PyTorch \cite{paszke2019pytorch}. We use regular sparse convolutions on the encoder blocks for a wider exchange of features and submanifold convolutions elsewhere for more efficient computation due to avoiding feature dilation. The encoder block uses a stride of 2. Each block uses a kernel size of $3$x$3$x$3$, except for the first sub-manifold convolution and the fully connected blocks, which use a kernel size of $1$x$1$x$1$. 

The architecture comprises a UNet backbone \cite{ronneberger2015u} with residual connections, followed by two smaller fully connected networks to extract the radii and directions. A ReLU activation function and batch normalization follow each convolutional layer. We add a fully-connected network head when branch-foliage segmentation is required. 

\begin{figure*}[hbtp]
\centering
\includegraphics[width=0.8\textwidth]{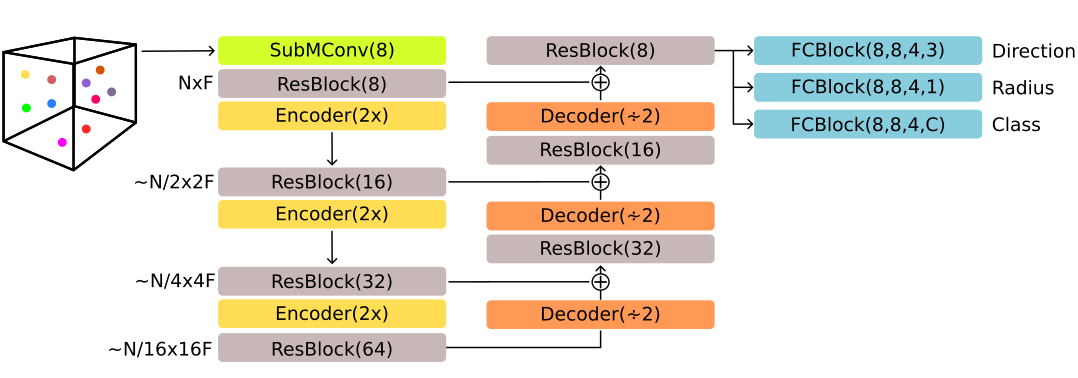}
\caption{Network architecture diagram.}
\label{fig:model}
\end{figure*}

A high-level overview of the network architecture is shown in Figure~\ref{fig:model}. 

A block sampling scheme ensures the network can process larger trees. During training, for each point cloud, we randomly sample (at each epoch) a $4m^{3}$ block and mask the outer regions of the block to avoid inaccurate predictions from the edges. We tile the blocks during inference, overlapping the masked regions.

We estimate a logarithmic radius to handle the variation in branch radii, which spans several orders of magnitude \cite{dassot2019assessing}, which provides a relative error. The loss function (Equation \ref{Loss_Function}) comprises two components. Firstly we use the L1-loss for the radius (Equation \ref{Radius_Loss}) and the cosine similarity (Equation \ref{Direction_Loss}) for the direction loss.
We use the Adam optimizer, a batch size of $16$, and a learning rate of $0.1$. The learning rate decays by a factor of $10$ if the validation data-set loss does not improve for $10$ consecutive epochs. 

\begin{tabular}{p{5cm}p{5cm}}
\begin{equation} \label{Direction_Loss}
L_{D} = \sum_{i = 0}^n \frac{Di \cdot \hat{Di}}{||Di||_2\cdot ||\hat{Di}||_2} 
\end{equation}
  &
\begin{equation} \label{Radius_Loss}
L_{R} = {\sum_{i = 0}^n |\ln(Ri) - \hat{Ri}|}
\end{equation} 
\end{tabular}

\begin{equation} \label{Loss_Function}
Loss = L_{R} + L_{D}
\end{equation} 

\subsection{Skeletonization Algorithm}
\vspace{-0.75cm}

\begin{figure}[H]
\centering
\subfloat[]{
  \includegraphics[width=0.16\textwidth, trim={5cm 1cm 2cm 5cm},clip]{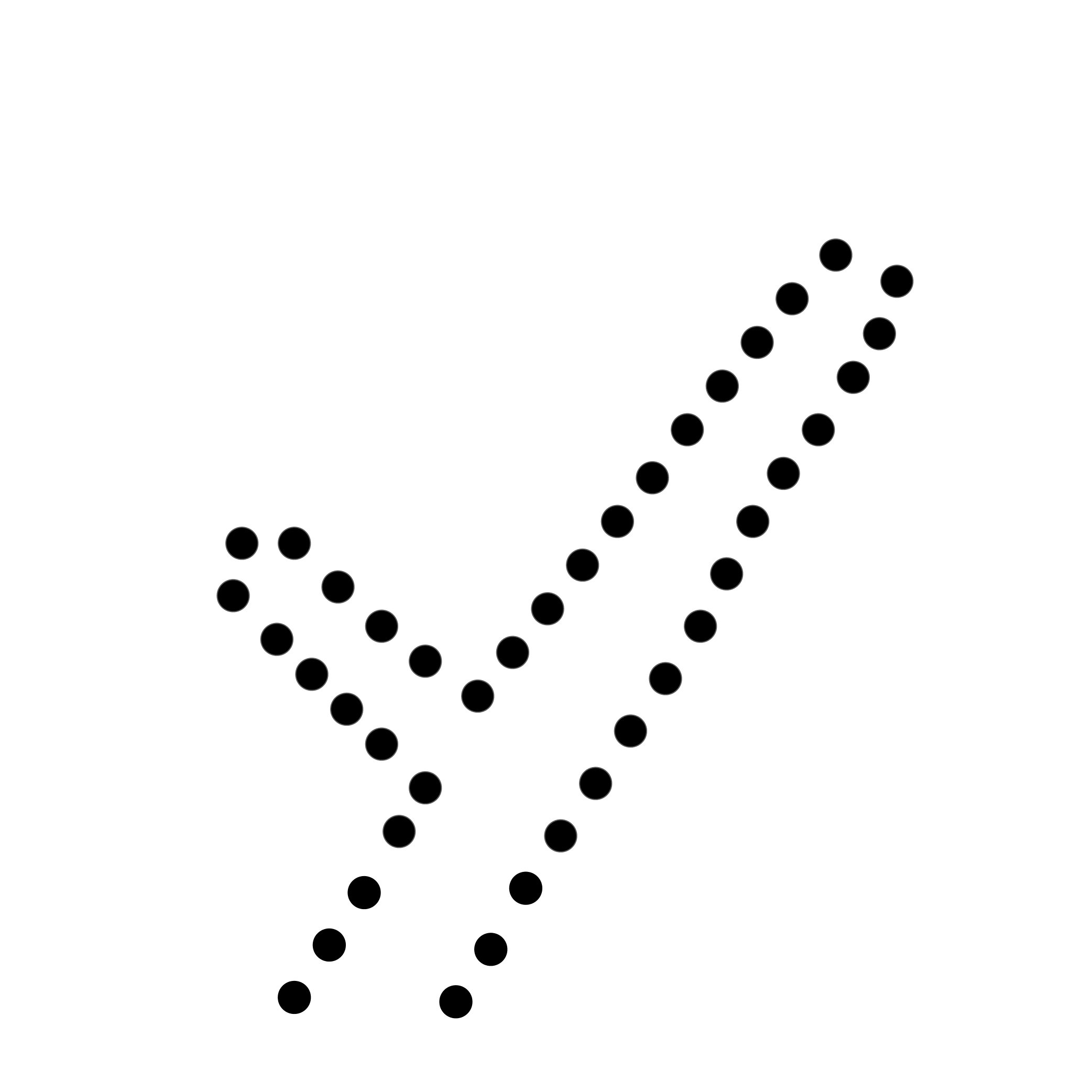}
  \label{fig:pipeline_a}
}
\subfloat[]{
  \includegraphics[width=0.16\textwidth, trim={5cm 1cm 2cm 5cm},clip]{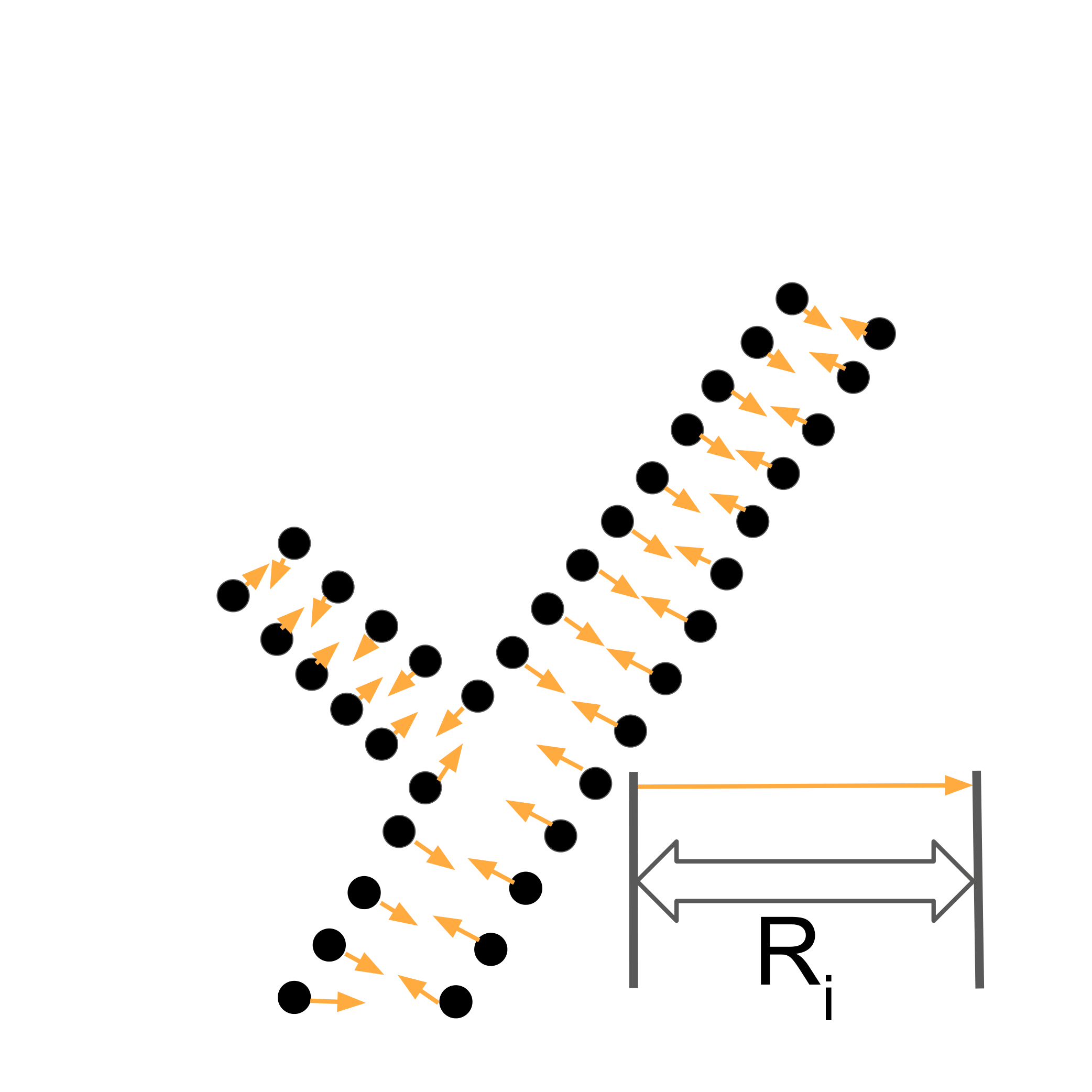}
  \label{fig:pipeline_b}
}
\subfloat[]{
  \includegraphics[width=0.16\textwidth, trim={5cm 1cm 2cm 5cm},clip]{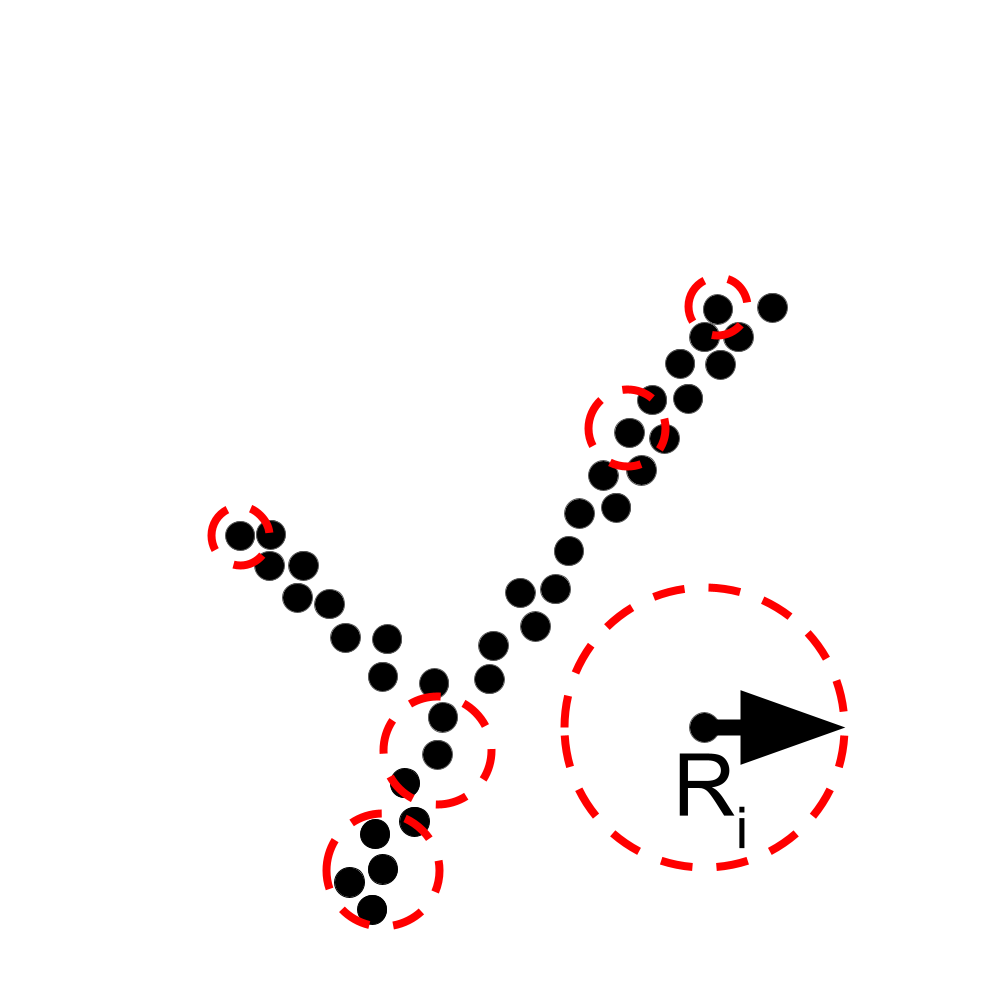}
  \label{fig:pipeline_c}
}
\subfloat[]{
  \includegraphics[width=0.16\textwidth, trim={5cm 1cm 2cm 5cm},clip]{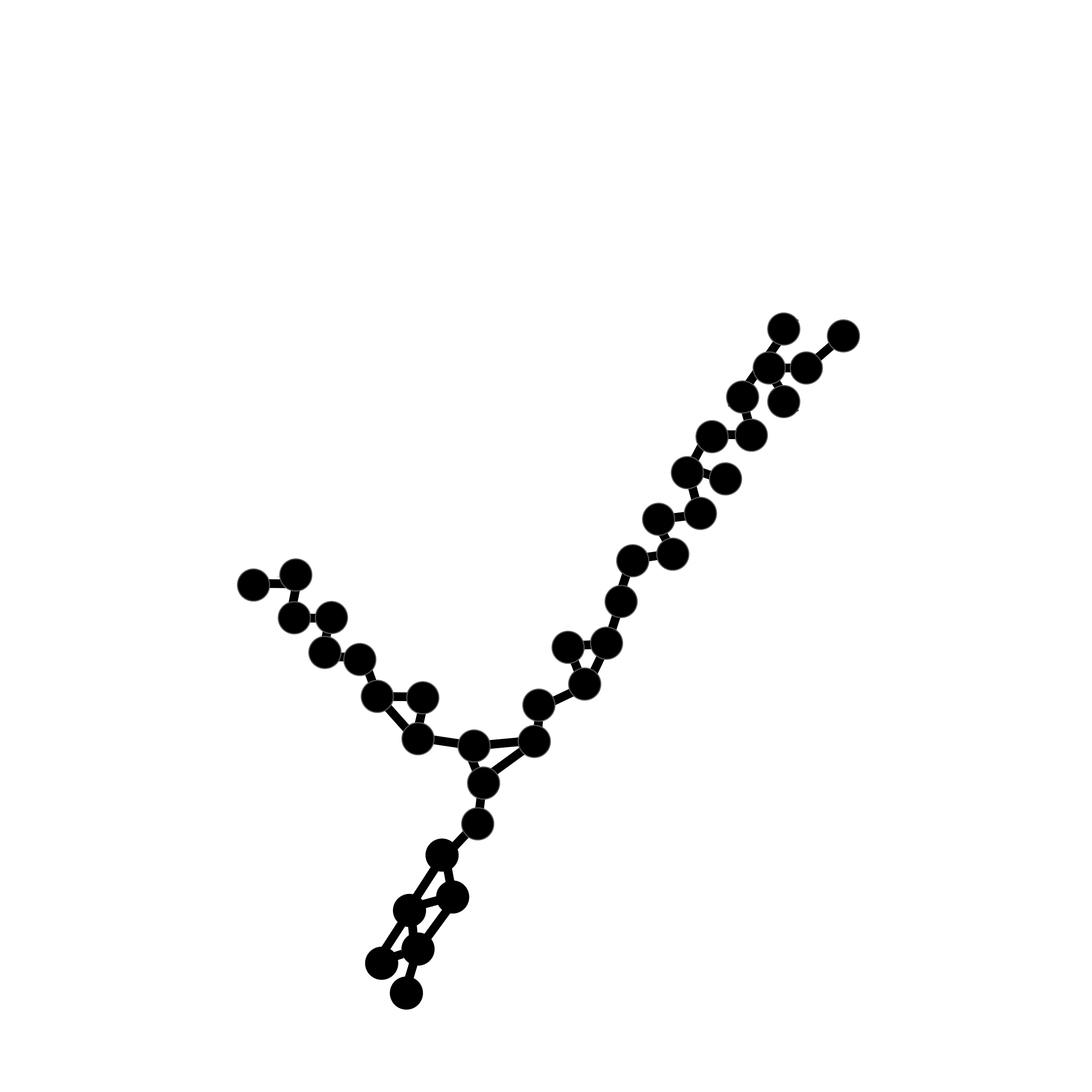}
  \label{fig:pipeline_d}
}
\hspace{0mm}
\\[-3ex]

\subfloat[]{
  \includegraphics[width=0.16\textwidth, trim={5cm 1cm 2cm 5cm},clip]{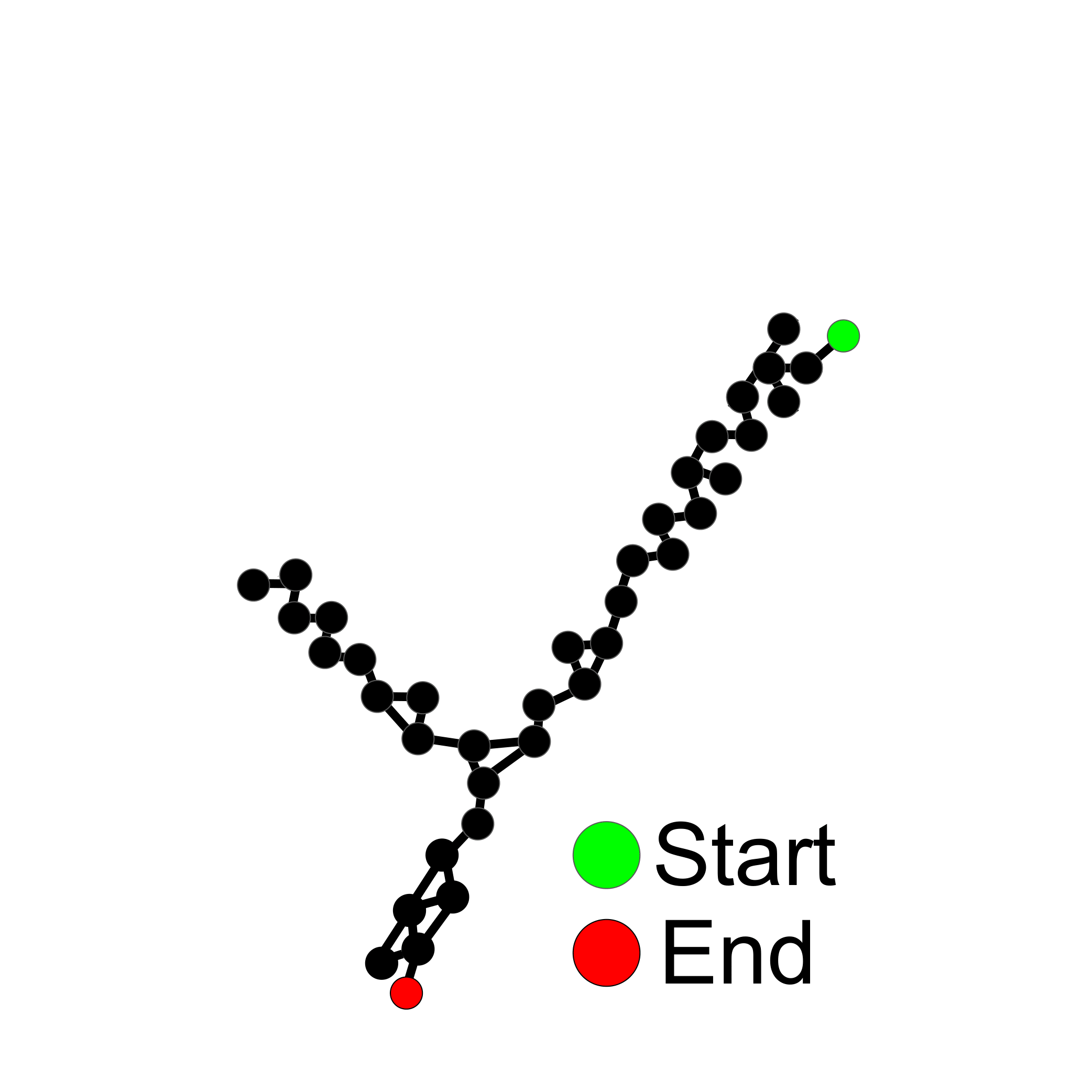}
  \label{fig:pipeline_e}
}
\subfloat[]{
  \includegraphics[width=0.16\textwidth, trim={5cm 1cm 2cm 5cm},clip]{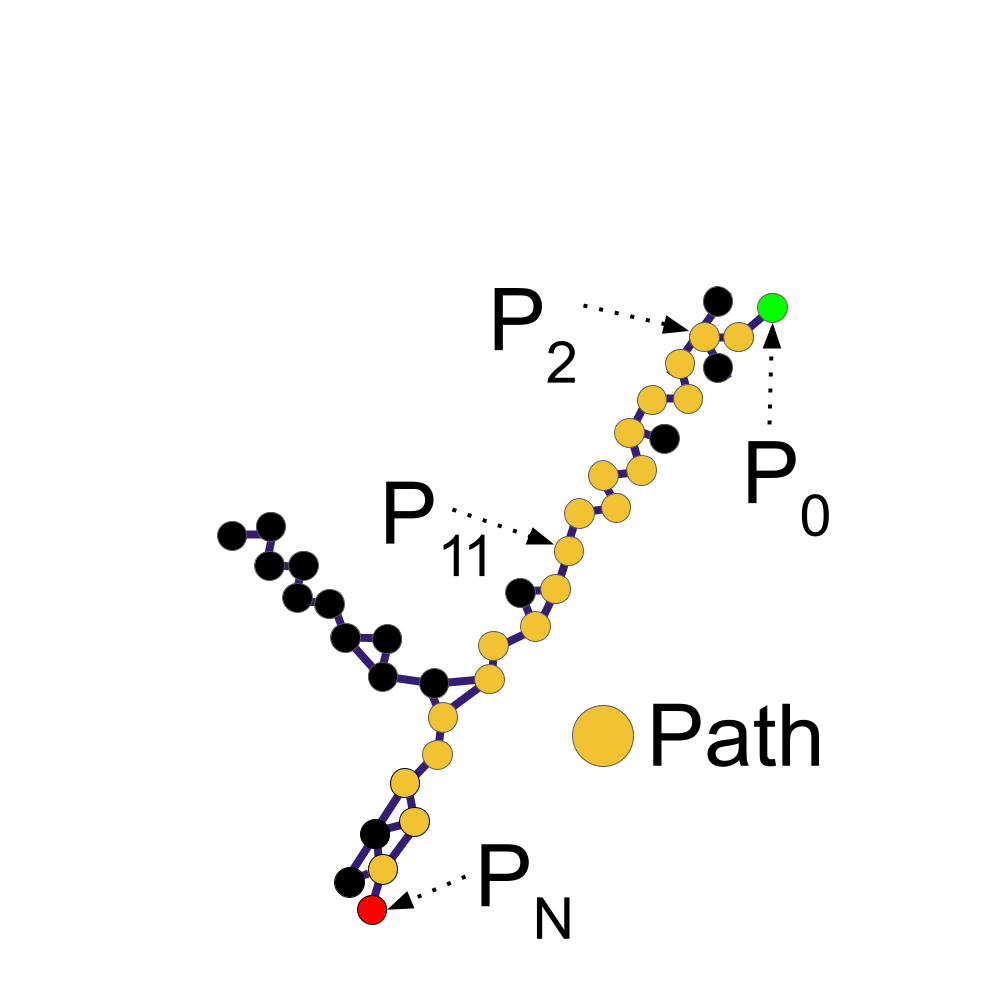}
  \label{fig:pipeline_f}
}
\subfloat[]{
  \includegraphics[width=0.16\textwidth, trim={5cm 1cm 2cm 5cm},clip]{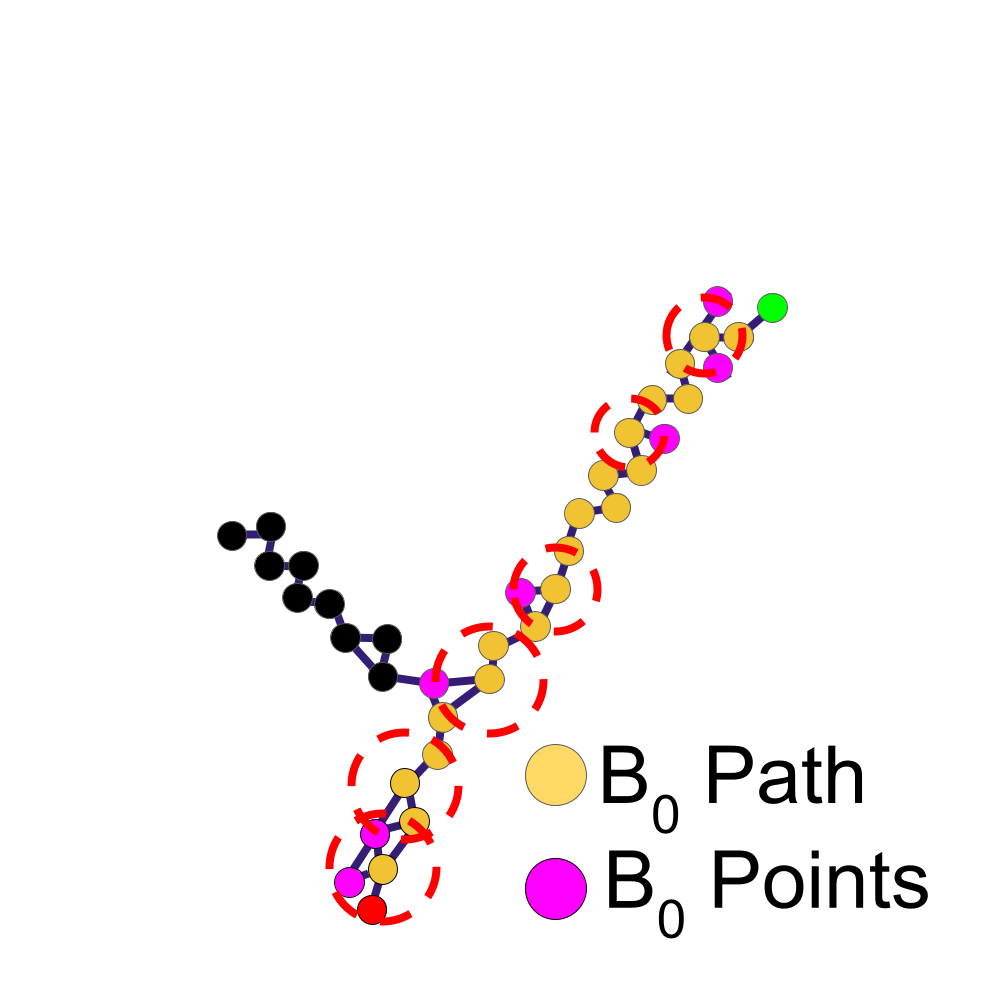}
  \label{fig:pipeline_g}
}
\subfloat[]{
  \includegraphics[width=0.16\textwidth, trim={5cm 1cm 2cm 5cm},clip]{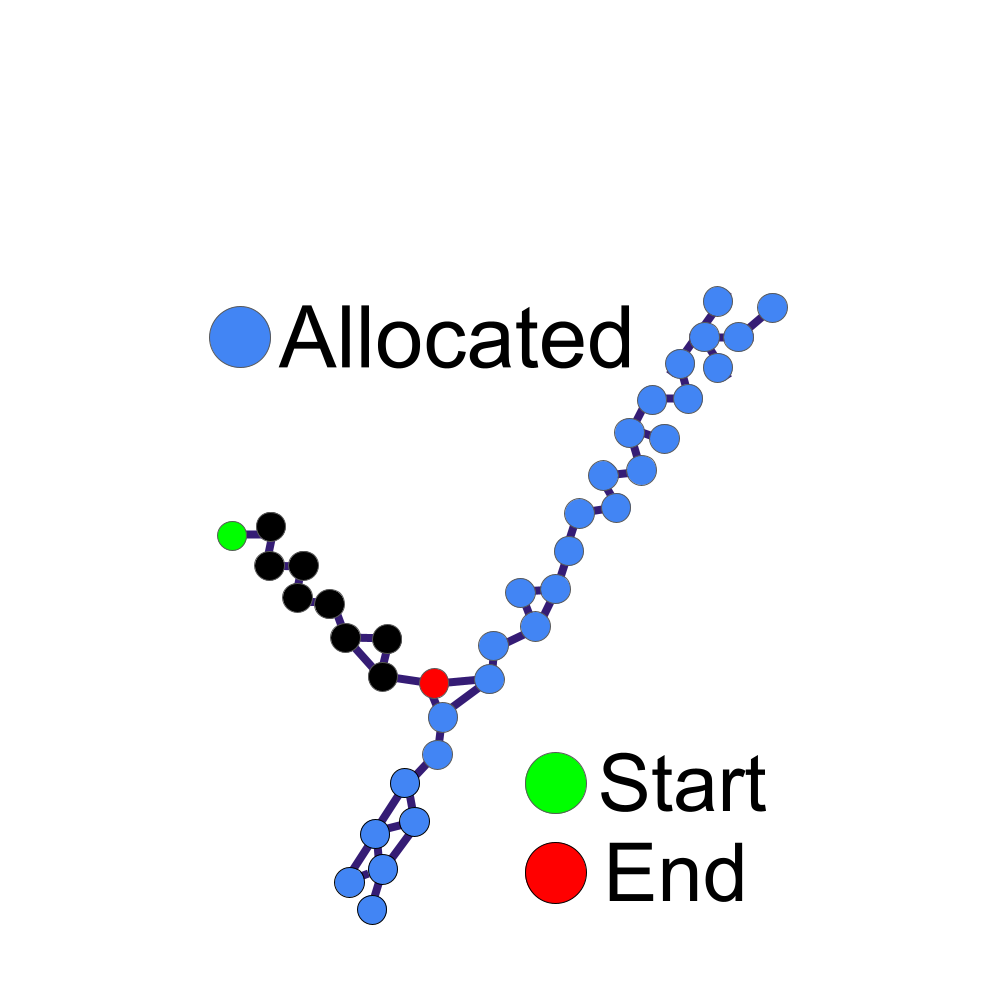}
  \label{fig:pipeline_h}
}
\hspace{0mm}
\\[-3ex]
\subfloat[]{
  \includegraphics[width=0.16\textwidth, trim={5cm 1cm 2cm 5cm},clip]{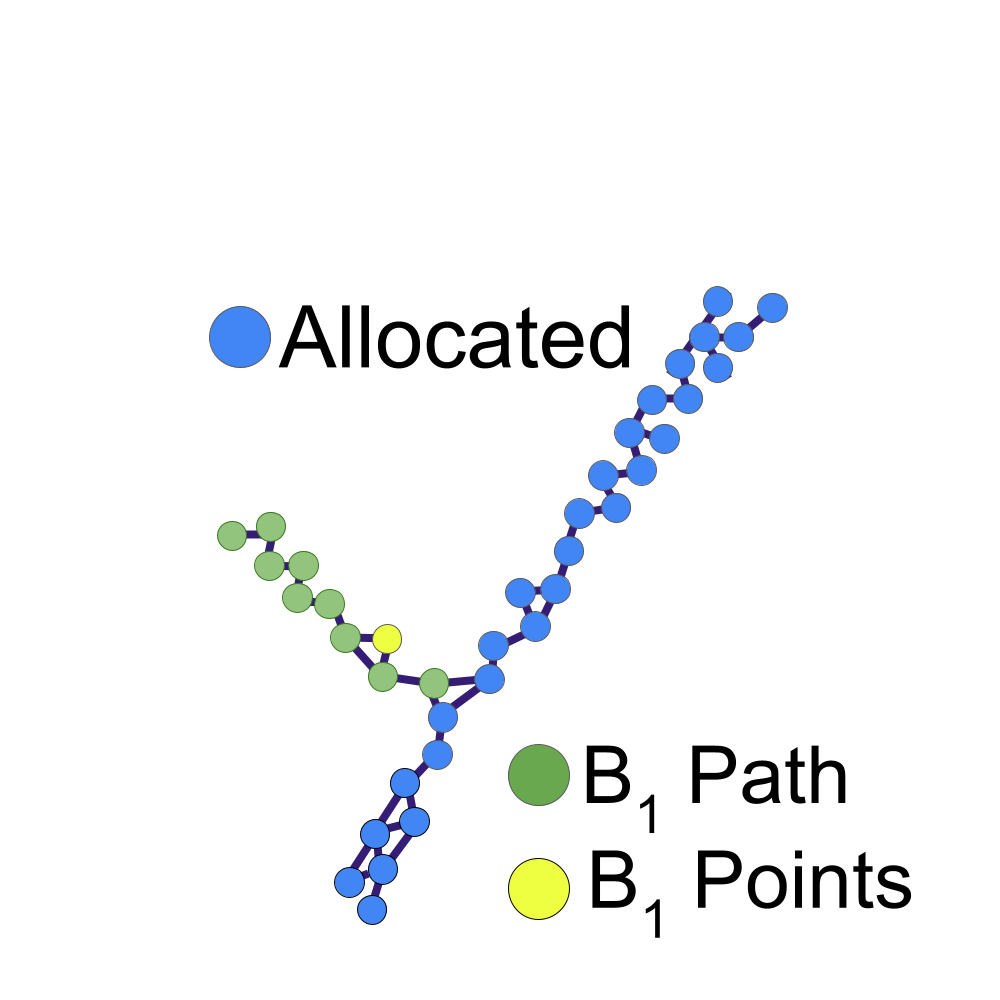}
  \label{fig:pipeline_i}
}
\subfloat[]{
  \includegraphics[width=0.16\textwidth, trim={5cm 1cm 2cm 5cm},clip]{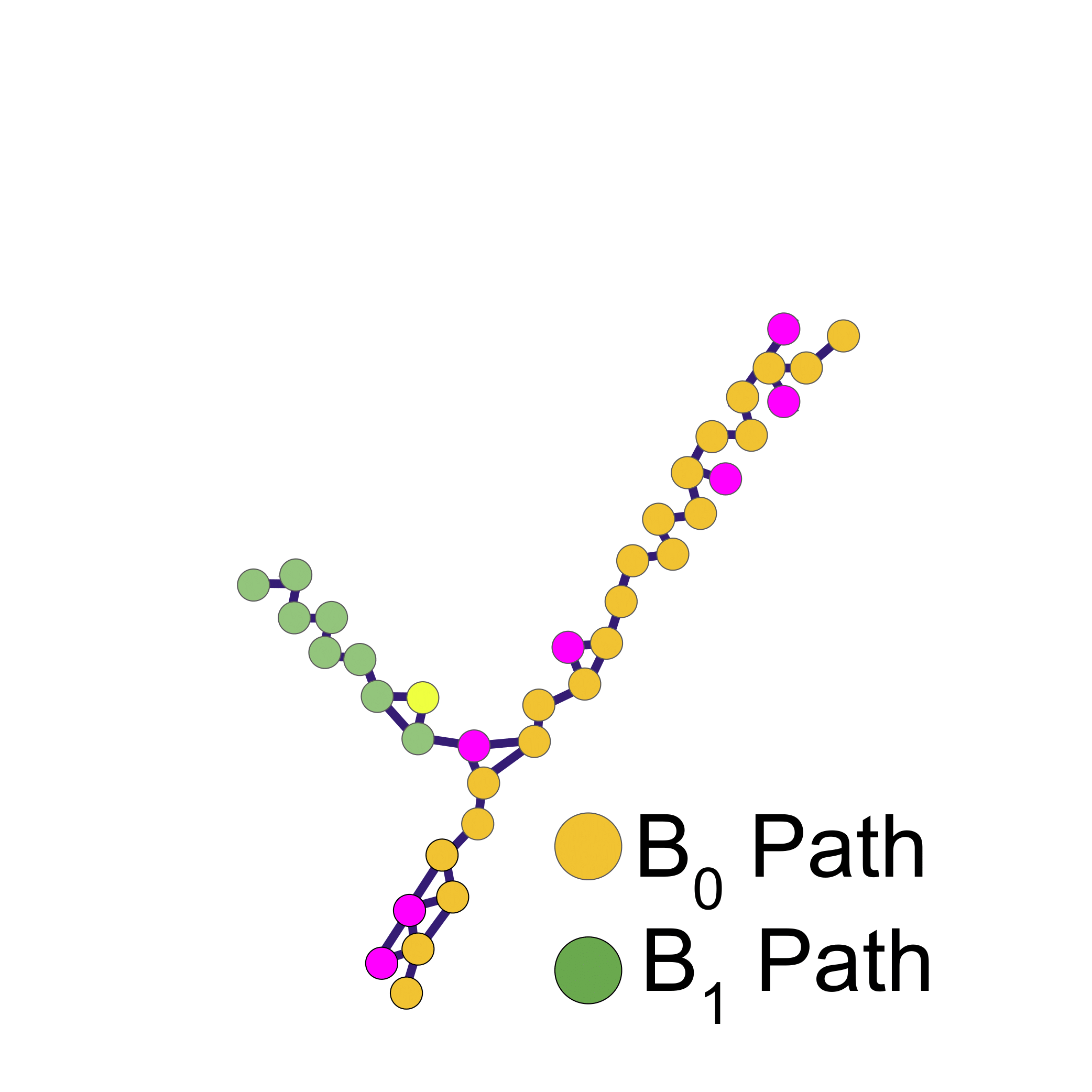}
  \label{fig:pipeline_j}
}
\subfloat[]{
  \includegraphics[width=0.16\textwidth, trim={5cm 1cm 0cm 5cm},clip]{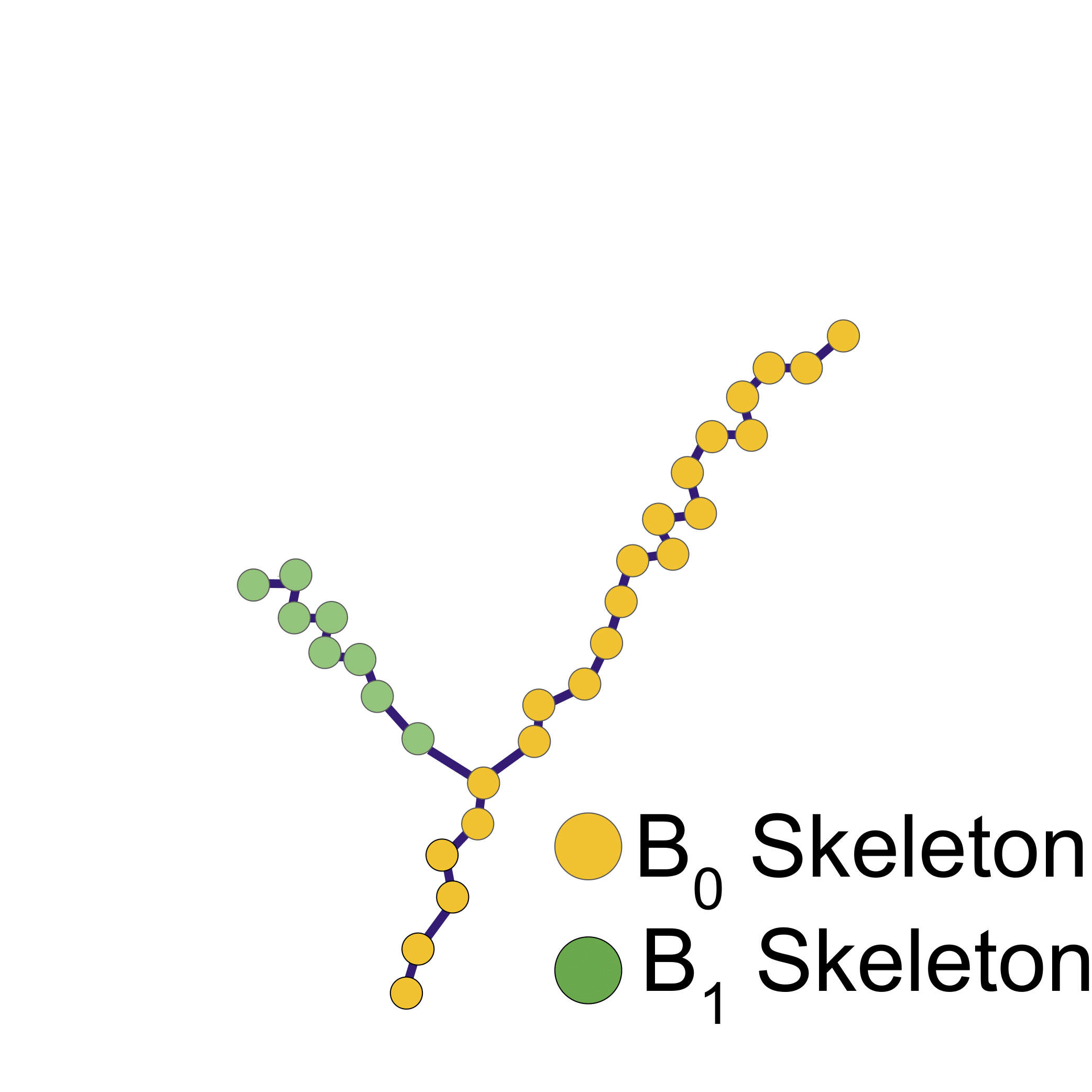}
  \label{fig:pipeline_l}

}
\subfloat[]{
  \includegraphics[width=0.16\textwidth, trim={5cm 1cm 0cm 5cm},clip]{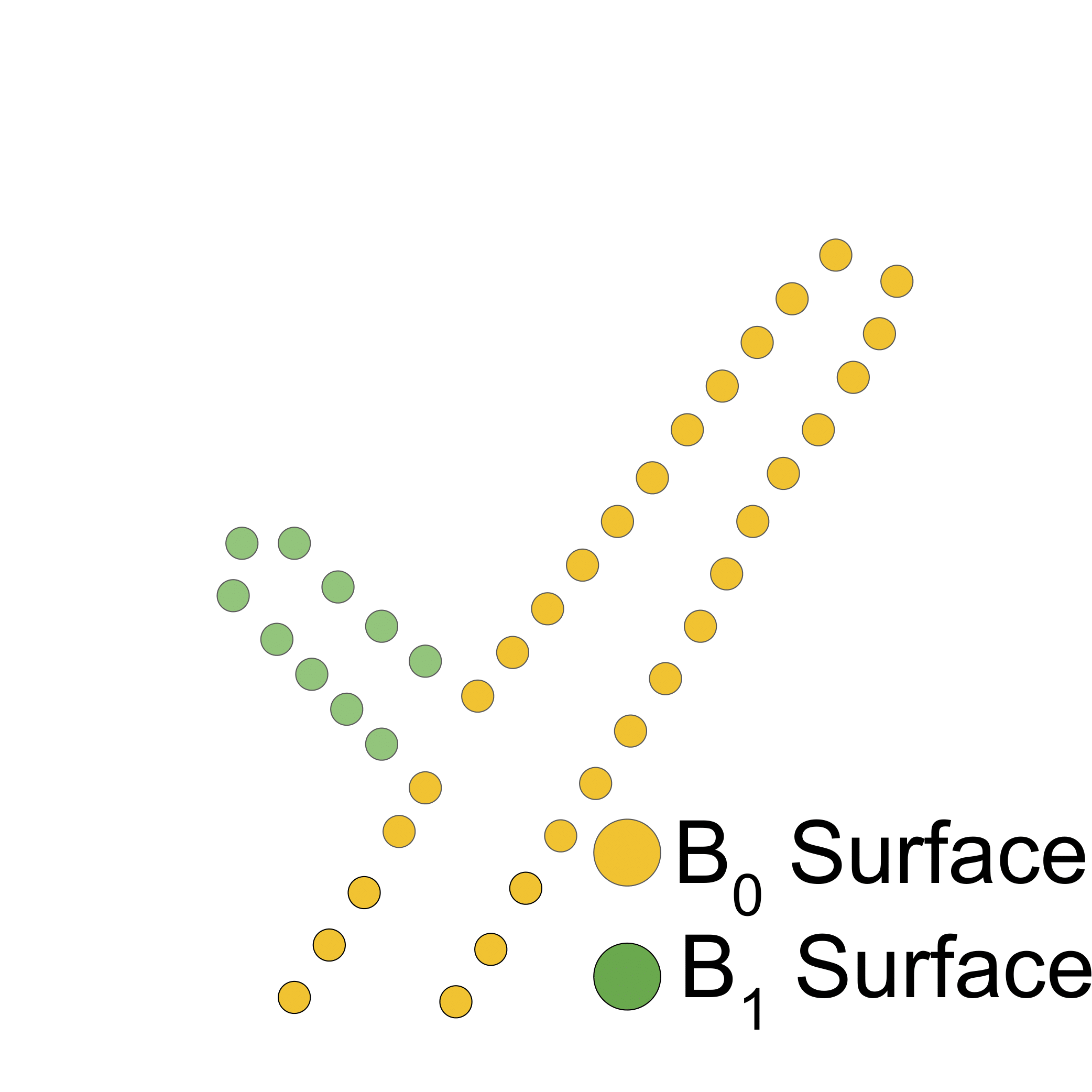}
  \label{fig:pipeline_m}
}
\caption{Skeletonization Algorithm: (a) Input points, (b) Medial axis approximation, (c) Neighbourhood radius search, (d) Neighbourhood graph, (e) $B_0$ Farthest point, (f) $B_0$ Trace path,(g) $B_0$ Allocated points, (h) $B_1$ Farthest (unallocated) point, (i) $B_1$ Trace path and allocated points, (k)  Branch skeletons, (i) Corresponding surface points.}
\label{fig:pipeline}

\end{figure}

Once $\left\{Ri\right\}$ and $\left\{Di\right\}$, have been predicted by the network. We use this information to project the input surface points $\left\{Pi\right\}$ (\ref{fig:pipeline_a}) onto the medial axis  (\ref{fig:pipeline_b}).

We form a neighbourhood graph  (\ref{fig:pipeline_d}) where points are connected to neighbours with weight equal to the distance between points and restricted to edges with a distance less than the predicted radius (\ref{fig:pipeline_c}). 

As the point cloud has gaps due to self-occlusion and noise, we end up with multiple connected components. Each connected component we call a sub-graph. We process each sub-graph sequentially. For each sub-graph:

\begin{enumerate}
    \item A distance tree is created based on the distance from the root node (the lowest point in each sub-graph - shown in red in (\ref{fig:pipeline_e})) to each point in the sub-graph.
    \item We assign each point a distance based on a Single Source Shortest Path  (SSSP) algorithm. A greedy algorithm extracts paths individually until all points are marked as allocated (steps $e$ to $j$).
    \item  We select a path to the furthest unallocated point and trace its path back to either the root (\ref{fig:pipeline_e}) or an allocated point (\ref{fig:pipeline_i}). 
    \item We add this path to a skeleton tree (\ref{fig:pipeline_f}).
    \item We mark points as allocated which lie within the predicted radius of the path (\ref{fig:pipeline_g}).
    \item We repeat this process until all points are allocated (\ref{fig:pipeline_i}, \ref{fig:pipeline_j})
\end{enumerate}

\section{Results} \label{results_section}
Performance evaluation of skeletonization algorithms is incredibly challenging and remains an open problem. Hence, there is no widely accepted metric used for evaluation. We compare our algorithms skeleton for quantitative evaluation using a form of precision and recall which matches points along the ground truth skeleton against points along the estimated skeleton.

We evaluate our method against the state-of-the-art AdTree algorithm \cite{du2019adtree}. As our metrics do not evaluate topological errors directly, additional qualitative analysis is conducted by visually inspecting the algorithm outputs against the ground truth.

Due to augmentations, some of the finer branches may become excessively noisy or disappear. To ensure the metrics measure what is possible to reconstruct, we prune the ground truth skeleton and point cloud based on a branch radius and length threshold - respective to tree size. 

\subsection{Metrics}

We evaluate our skeletons using point cloud metrics by sampling our skeletons at a 1mm resolution.
We use the following metrics to assess our approach: f-score, precision, recall and AUC. 
For the following metrics, we consider \(p \in \mathcal{S} \) points along the ground truth skeleton and \(p^* \in \mathcal{S}^* \) points obtained by sampling the output skeleton. \( p_r \) is the radius at each point. 
We use a threshold variable \({t}\), which sets the distance points must be within based on a factor of the ground truth radius. We test this over the range of \(0.0 - 1.0\). The f-score is the harmonic mean of the precision and recall. 

 \subsubsection{Skeletonization Precision}
To calculate the precision, we first get the nearest points from the output skeleton $p_i \in \mathcal{S}$ to the ground truth skeleton $p_j^* \in \mathcal{S^*}$, using a distance metric of the euclidean distance relative to the ground truth radius $r_j^*$.
The operator $\llbracket . \rrbracket$ is the Iverson bracket, which evaluates to 1 when the condition is true; otherwise, 0.

\begin{equation}
\begin{split}
d_{ij} = ||p_i - p_j^*||
\end{split}
\end{equation}
\begin{equation}\label{precision}
\begin{split}
P(t) = \frac{100}{|S|} \sum_{i \in \mathcal{S}} \llbracket  d_{ij} < t \ r_j^* \land \mathop{\forall}_{k \in \mathcal{S}} d_{ij} \leq d_{kj} \rrbracket 
\end{split}
\end{equation}
\subsubsection{Skeletonization Recall}
To calculate the recall, we first get the nearest points from the ground truth skeleton $p_j^* \in \mathcal{S^*}$ to the output skeleton $p_i \in \mathcal{S}$. We then calculate which points fall inside the thresholded ground truth radius. This gives us a measurement of the completeness of the output skeleton.
 
%At most one point $p^*$ (included in the set of recall true positives ($Q$))  is allowed to match the same $p$ where the distance is less than %the threshold. 

\begin{equation}\label{recall_truth_positives}
\begin{split}
R(t) = \frac{100}{|S^*|} \sum_{j \in \mathcal{S^*}} \llbracket  d_{ij} < t \ r_j^* \land \mathop{\forall}_{k \in \mathcal{S^*}} d_{ij} \leq d_{ik} \rrbracket 
\end{split}
\end{equation}

\subsection{Quantitative Results}
\begin{figure}
\centering
\begin{tabular}{ccc}
  \includegraphics[width=40mm]{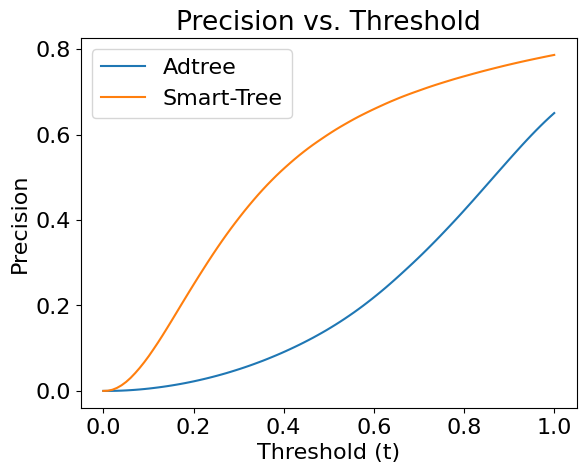} &   \includegraphics[width=40mm]{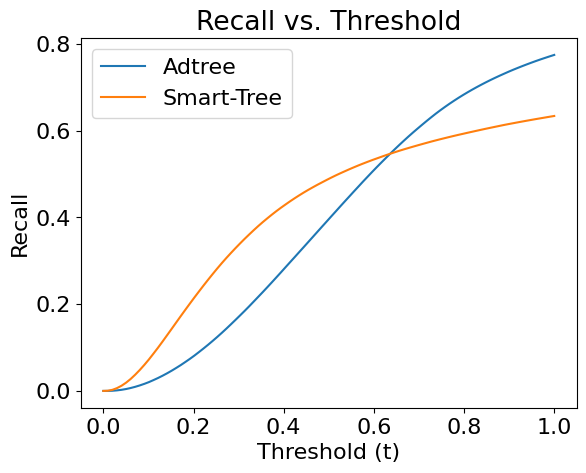}  &
 \includegraphics[width=40mm]{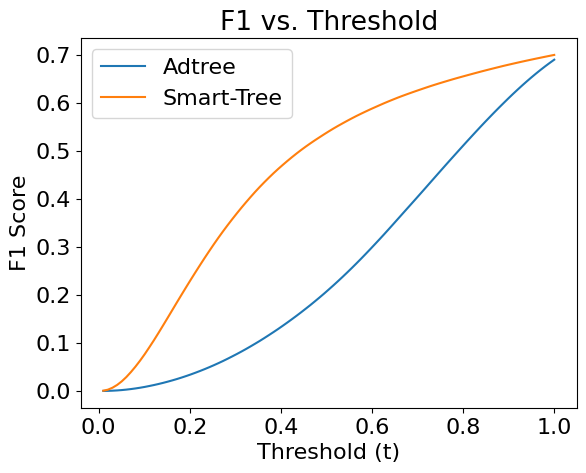} \\
\end{tabular}
\caption{Left to right: Precision Results (a), Recall Results (b), F1 Results (c).}
\label{fig:result-graphs}
\vspace{-2mm}%Put here to reduce too much white space after your table 
\end{figure}

We evaluate our method on our test set of sixty synthetic ground truth skeletons (made up of six species). Our results are summarized in Table \ref{table:results} and illustrated in Figure \ref{fig:result-graphs}. We compute the AUC for F1, precision, and recall using the radius threshold \({t}\) ranging from 0 to 1.
\begin{table}[htp]
\centering
\caption{Skeletonization Results.}
\begin{tabular*}{\textwidth}{
     P{0.30\textwidth} |                 
     P{0.30\textwidth} |                 
     P{0.30\textwidth} }
Metric & Smart-Tree & AdTree \\
\hline
Precision AUC & 0.53 & 0.21\\
Recall AUC & 0.40 & 0.38\\
F1 AUC & 0.45 & 0.26 
\end{tabular*}
\label{table:results}
\end{table}

Smart-Tree achieves a high precision score, with most points being close to the ground truth skeleton (Fig. \ref{fig:result-graphs}a). 
Compared to AdTree, Smart-Tree has lower recall at the most permissive thresholds, and this is due to Smart-Tree's inability to approximate missing regions of the point cloud, making it prone to gaps. AdTree, on the other hand, benefits from approximating missing regions. However, this also leads to AdTree having more topological errors and lower precision. Smart-Tree consistently achieves a higher F1 score and AUC for precision, recall, and F1, respectively. 
% We plan to curate an annotated set of real point clouds in the future, once we have acquired data from a range of species. 

% \begin{table}[htp ]
% \centering

% \caption{Skeletonization Results Across Thresholds.}

% \medskip
% \begin{tabular*}{\textwidth}{
%     P{0.12\textwidth}                   % Column 1
%     P{0.14\textwidth}                   % Column 1
%     P{0.14\textwidth}                   % Column 1
%     P{0.14\textwidth}                   % Column 1
%     P{0.14\textwidth}                   % Column 1
%     P{0.14\textwidth}                   % Column 1
%     P{0.14\textwidth}                   % Column 1
% }

% \toprule
% Threshold \({t}\)
% & \multicolumn{2}{c}{Precision Mean (\%)} 
% & \multicolumn{2}{c}{Recall Mean (\%)} 
% & \multicolumn{2}{c}{F1  Mean (\%)} \\
% \cmidrule(lr){2-3} \cmidrule(l){4-5} \cmidrule(l){6-7}
%  &  
% {AdTree} &   
% {Smart-Tree} & 
% {AdTree} &   
% {Smart-Tree} & 
% {AdTree} &
% {Smart-Tree}\\
% \midrule
% 0.1&0.54&9.75&1.98&6.73&0.89&7.80 \\
% \midrule
% 0.2&2.23&28.86&7.98&19.12&3.36&22.52 \\
% \midrule
% 0.3&5.05&45.77&17.06&29.27&7.53&34.95 \\
% \midrule
% 0.4&9.09&57.32&28.03&35.79&13.30&43.12 \\
% \midrule
% 0.5&14.54&64.64&39.58&39.78&20.70&48.15 \\
% \midrule
% 0.6&21.89&69.42&50.87&42.48&29.91&51.50 \\
% \midrule
% 0.7&32.27&72.89&60.72&44.64&40.40&54.10 \\
% \midrule
% 0.8&42.22&75.64&68.33&46.45&51.08&56.23 \\
% \bottomrule
% \label{table:results}
% \end{tabular*}
% \vspace{-15mm}%Put here to reduce too much white space after your table 
% \end{table}

\subsection{Qualitative Results}
\begin{figure}[H]
\centering
\begin{tabular}{cccc}
  \includegraphics[width=0.24\textwidth]{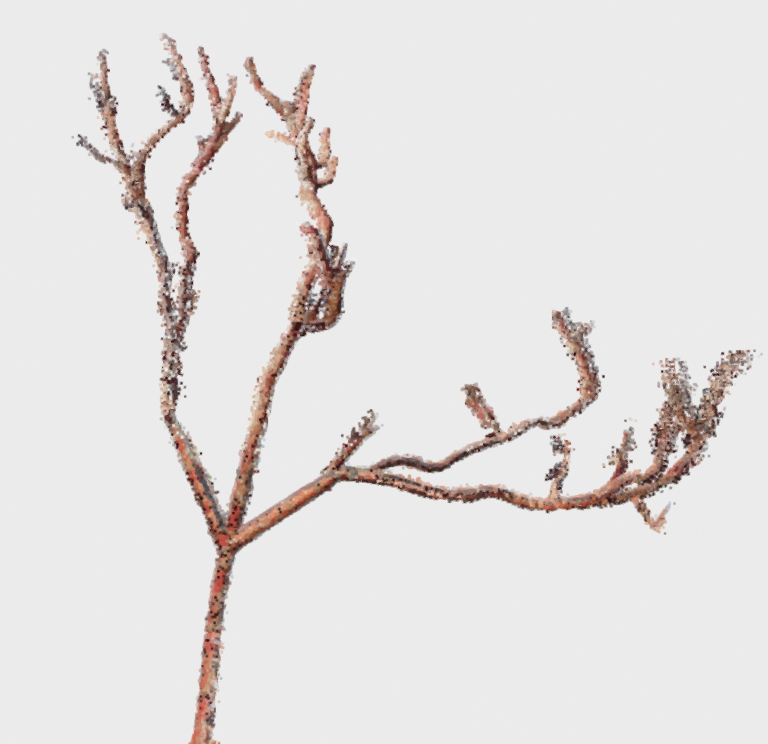} &  
  \includegraphics[width=0.24\textwidth]{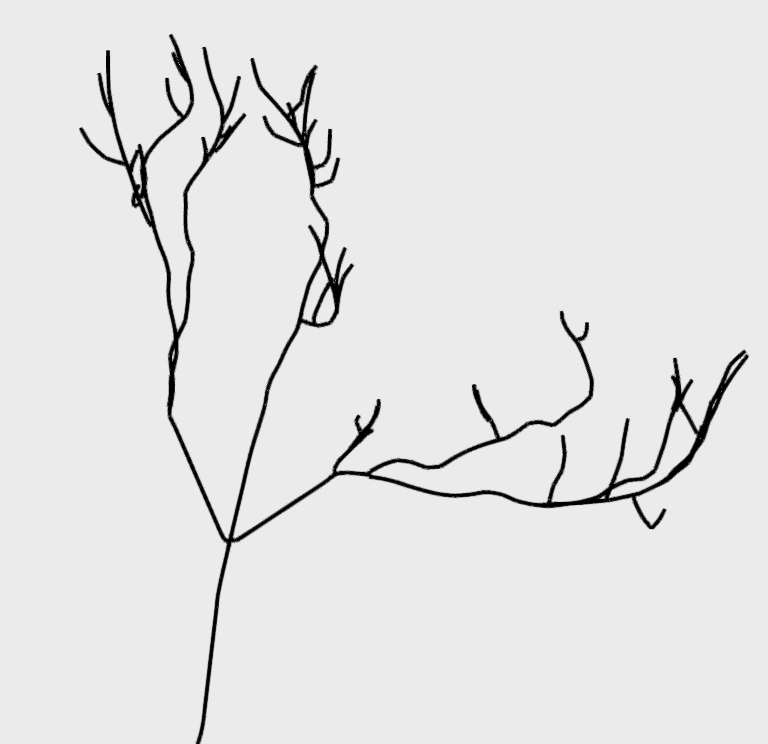}  &
  \includegraphics[width=0.24\textwidth]{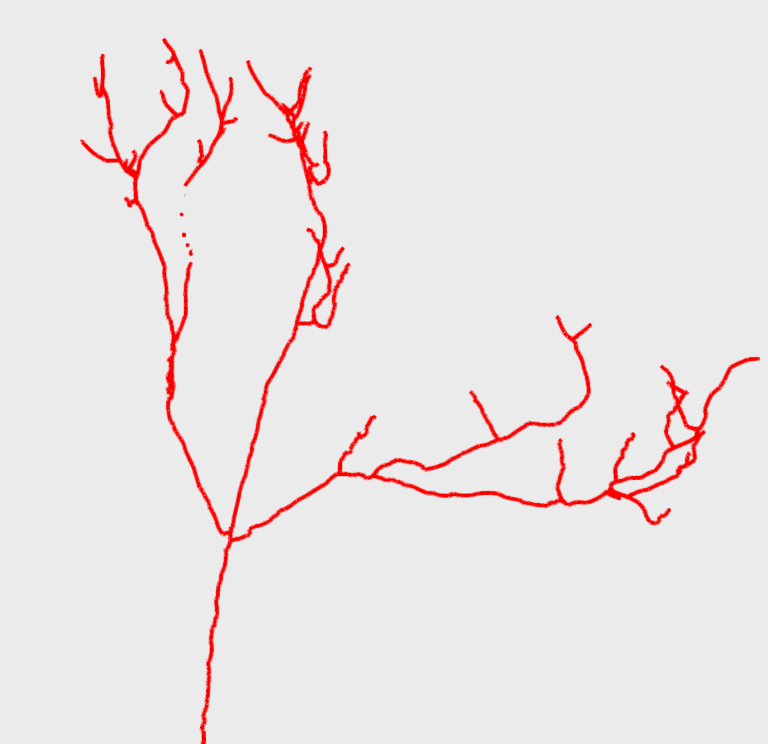}  &
  \includegraphics[width=0.24\textwidth]{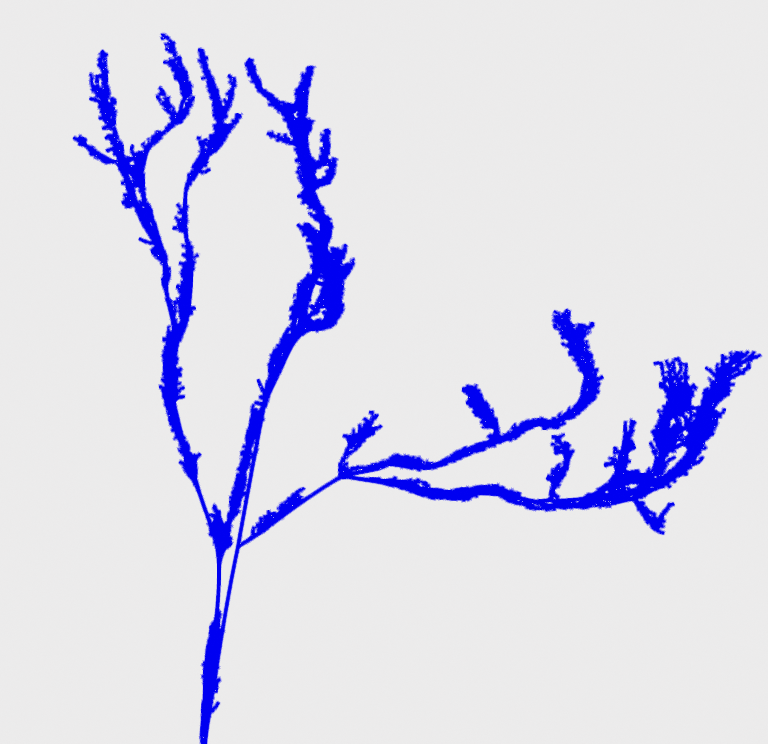}  \\
  \includegraphics[width=0.24\textwidth]{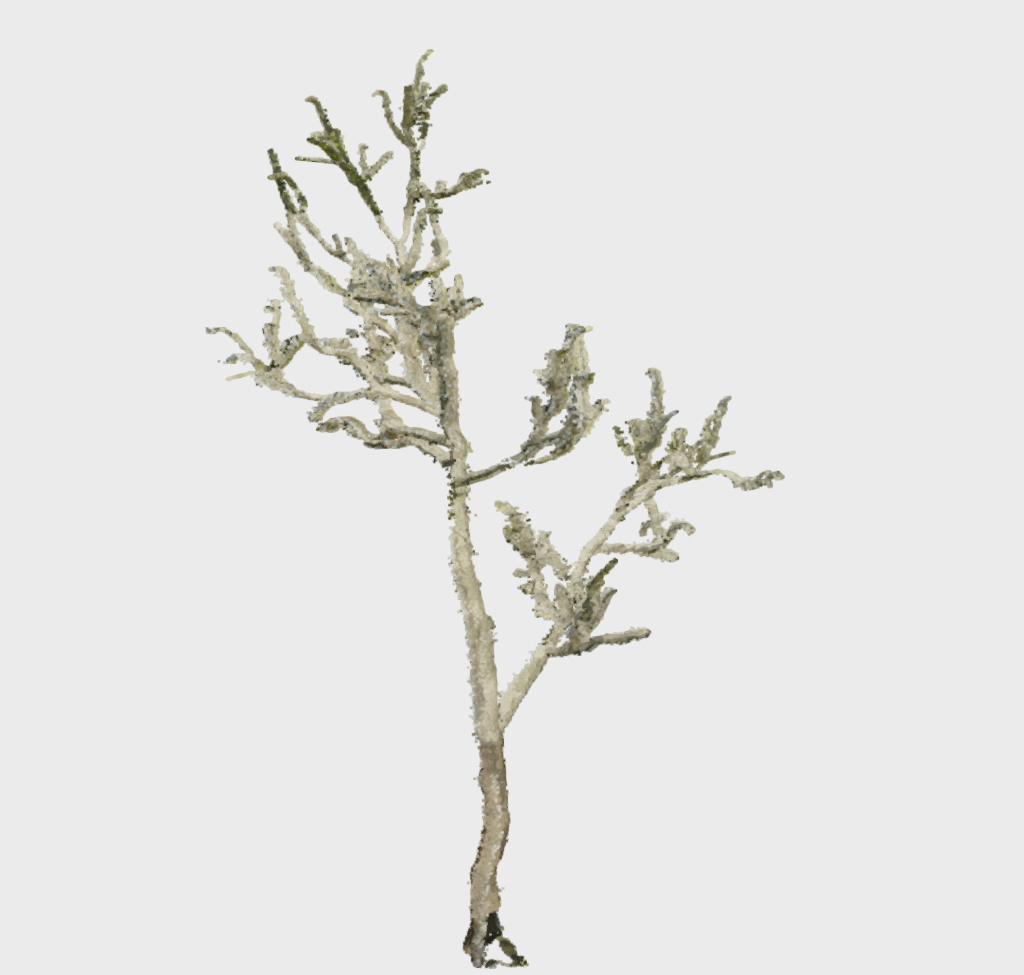} &  
  \includegraphics[width=0.24\textwidth]{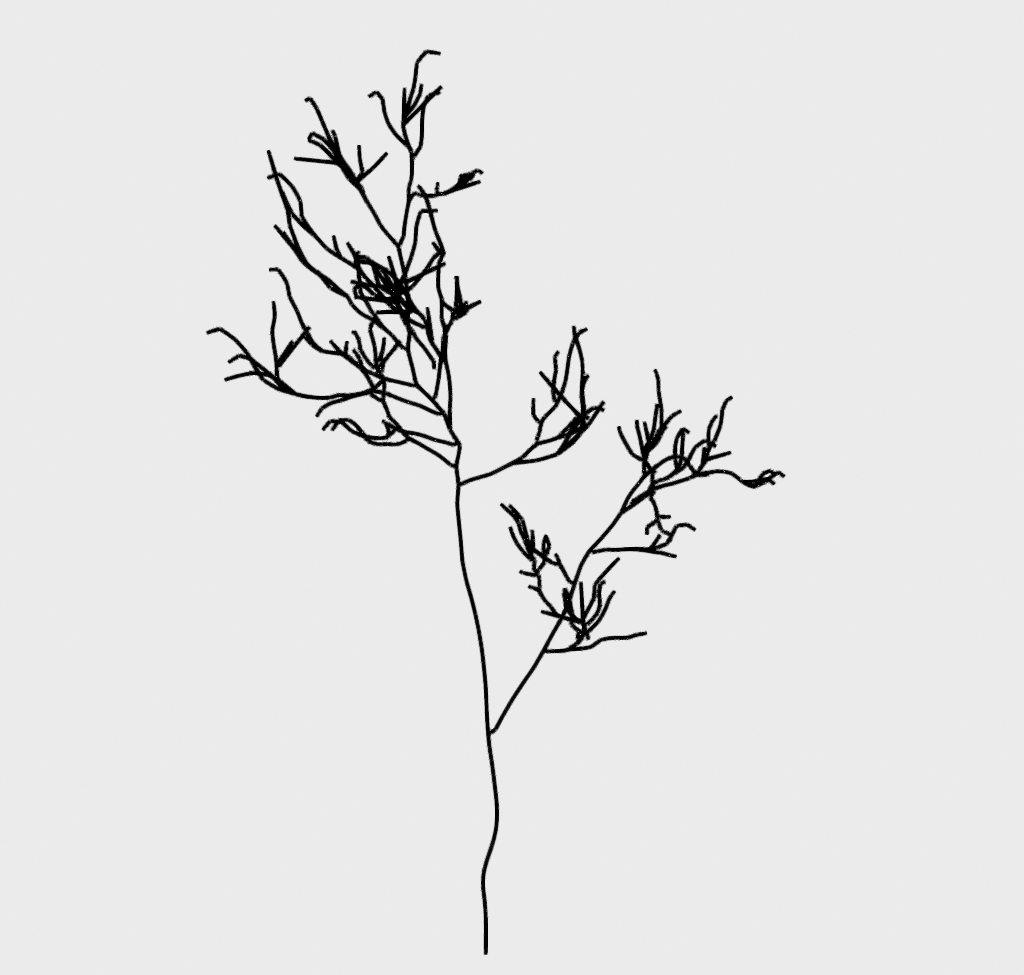}  &
  \includegraphics[width=0.24\textwidth]{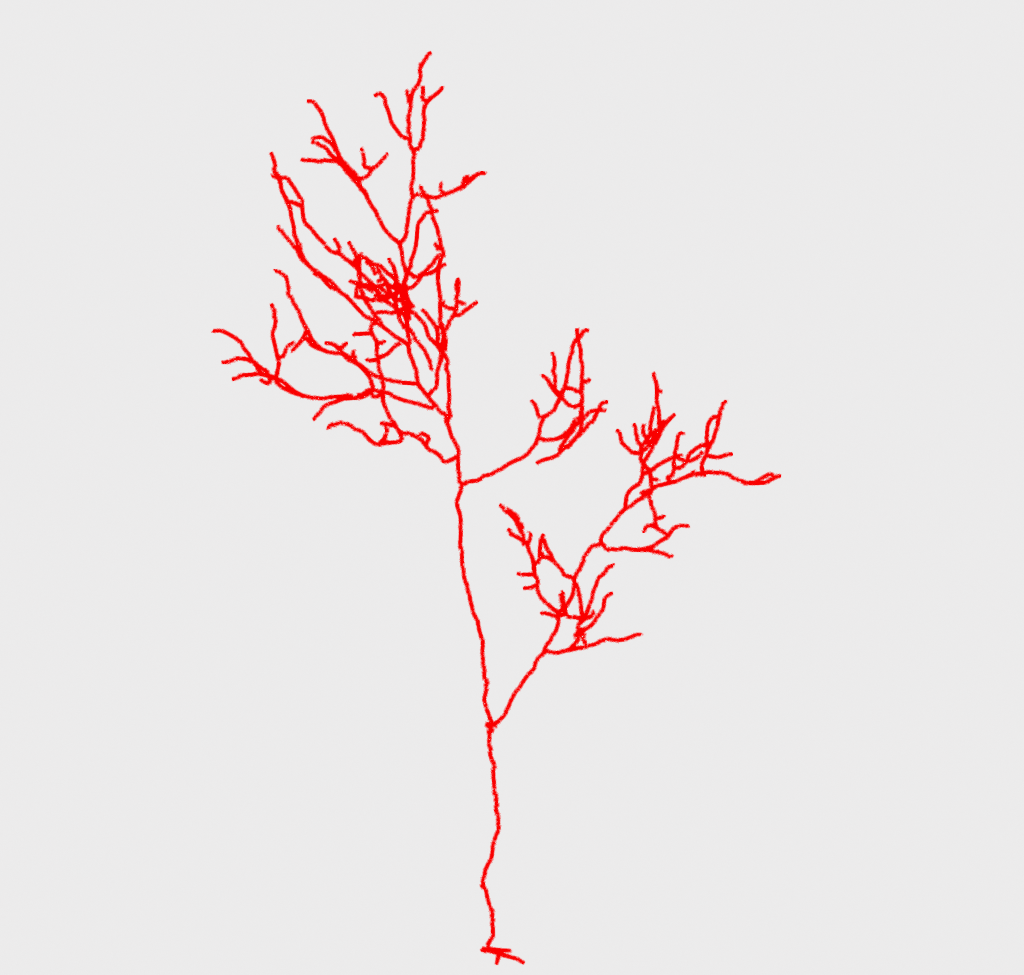}  &
  \includegraphics[width=0.24\textwidth]{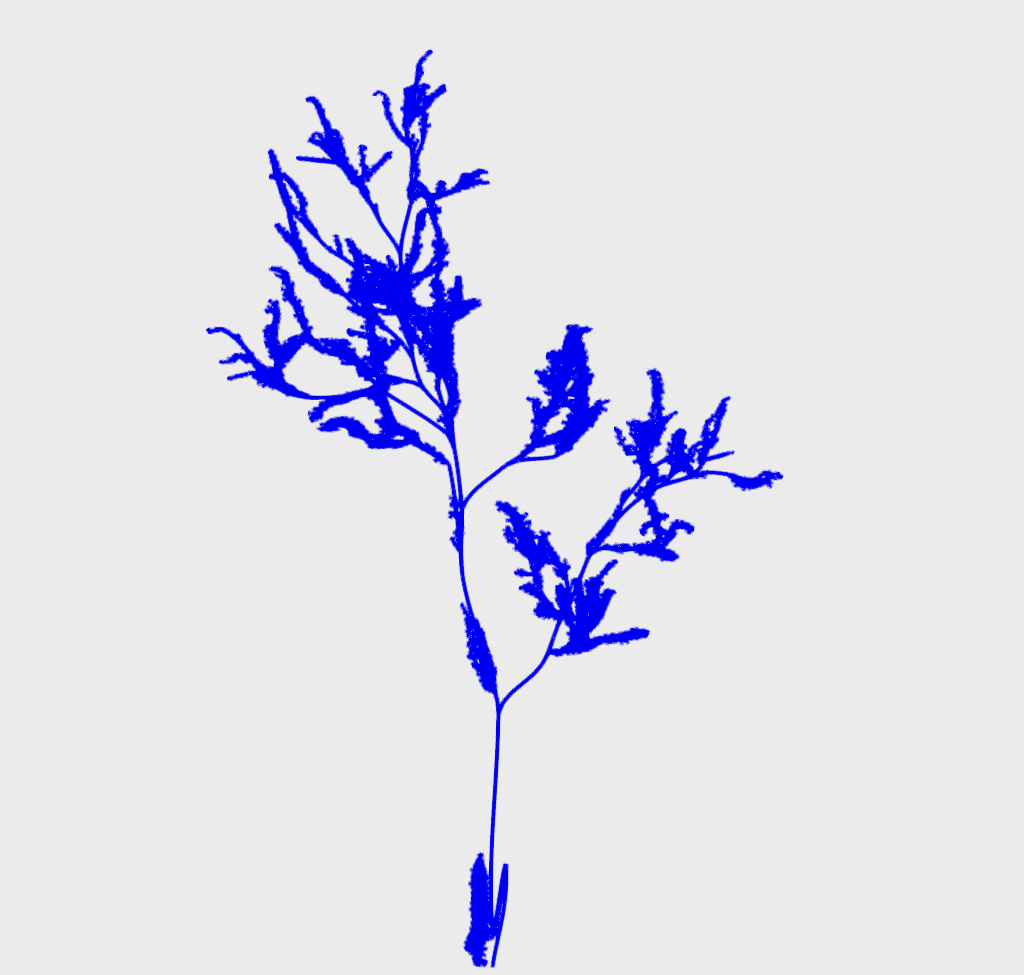}  \\
  \includegraphics[width=0.24\textwidth]{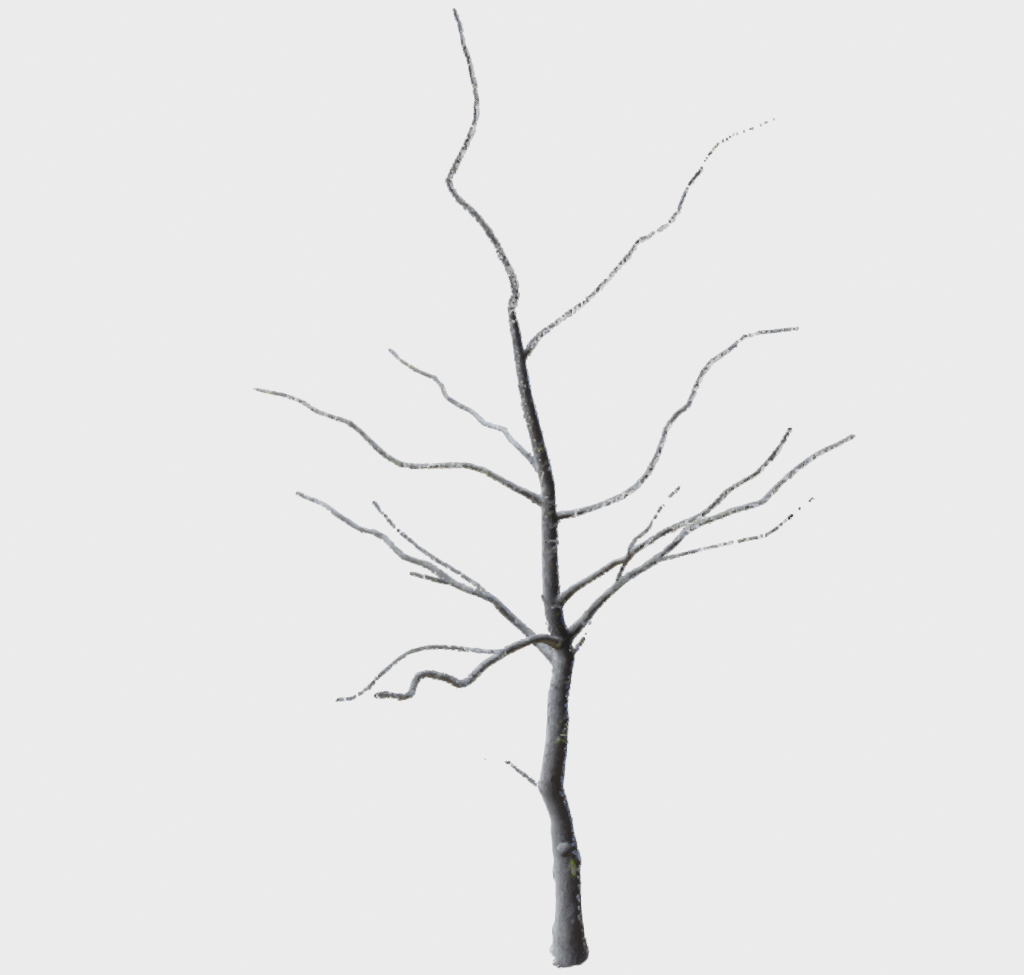} &  
  \includegraphics[width=0.24\textwidth]{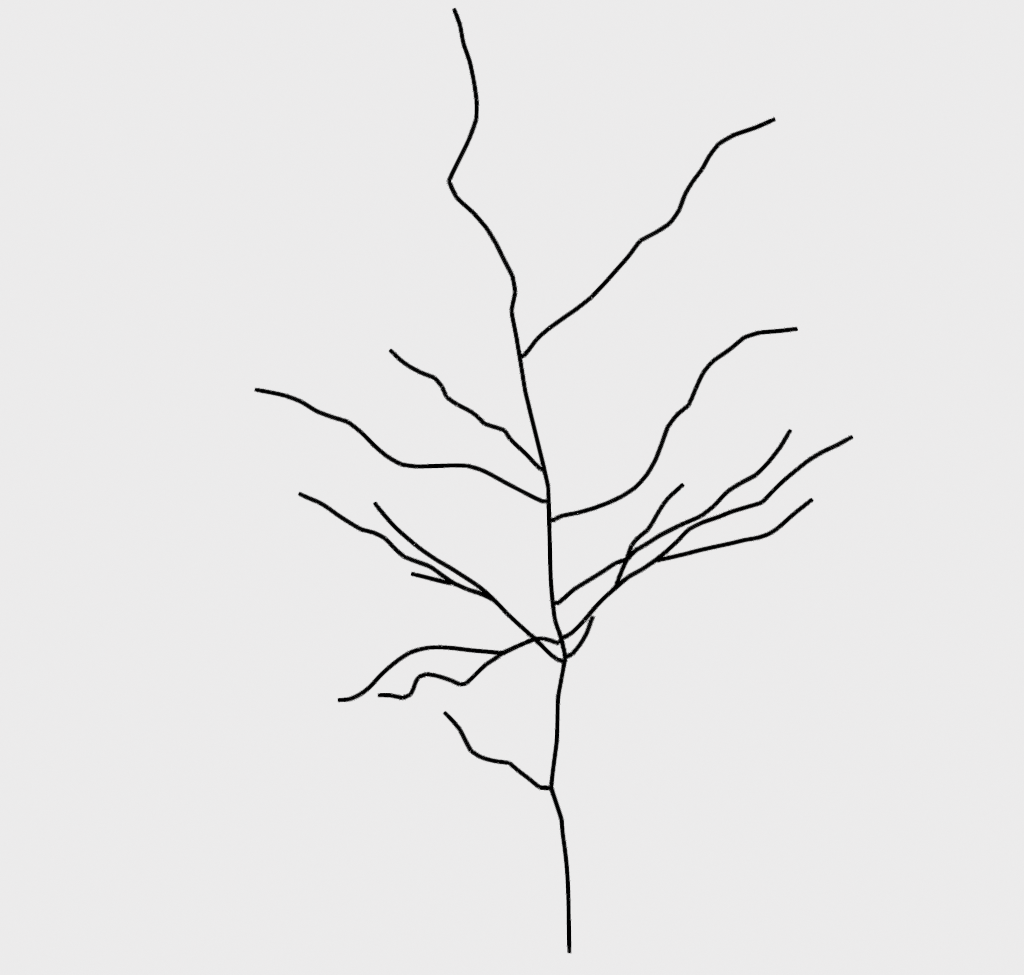}  &
  \includegraphics[width=0.24\textwidth]{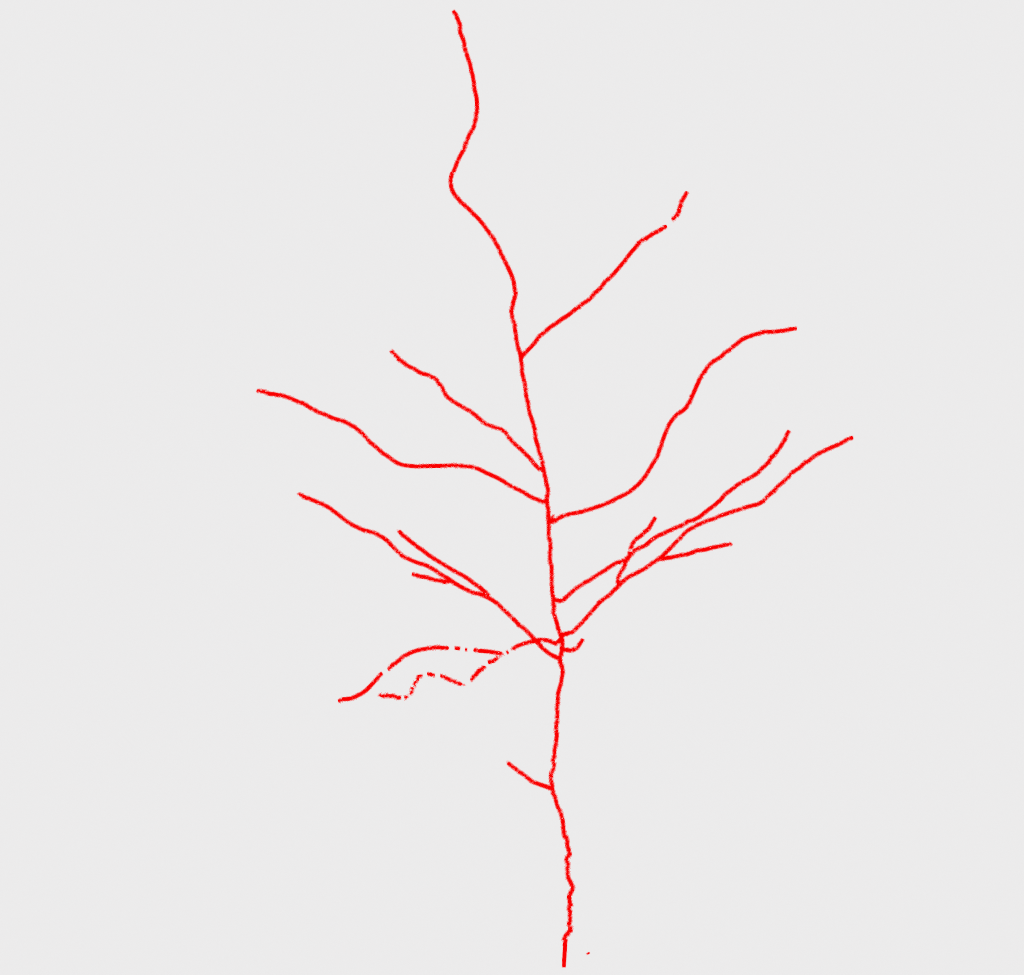}  &
  \includegraphics[width=0.24\textwidth]{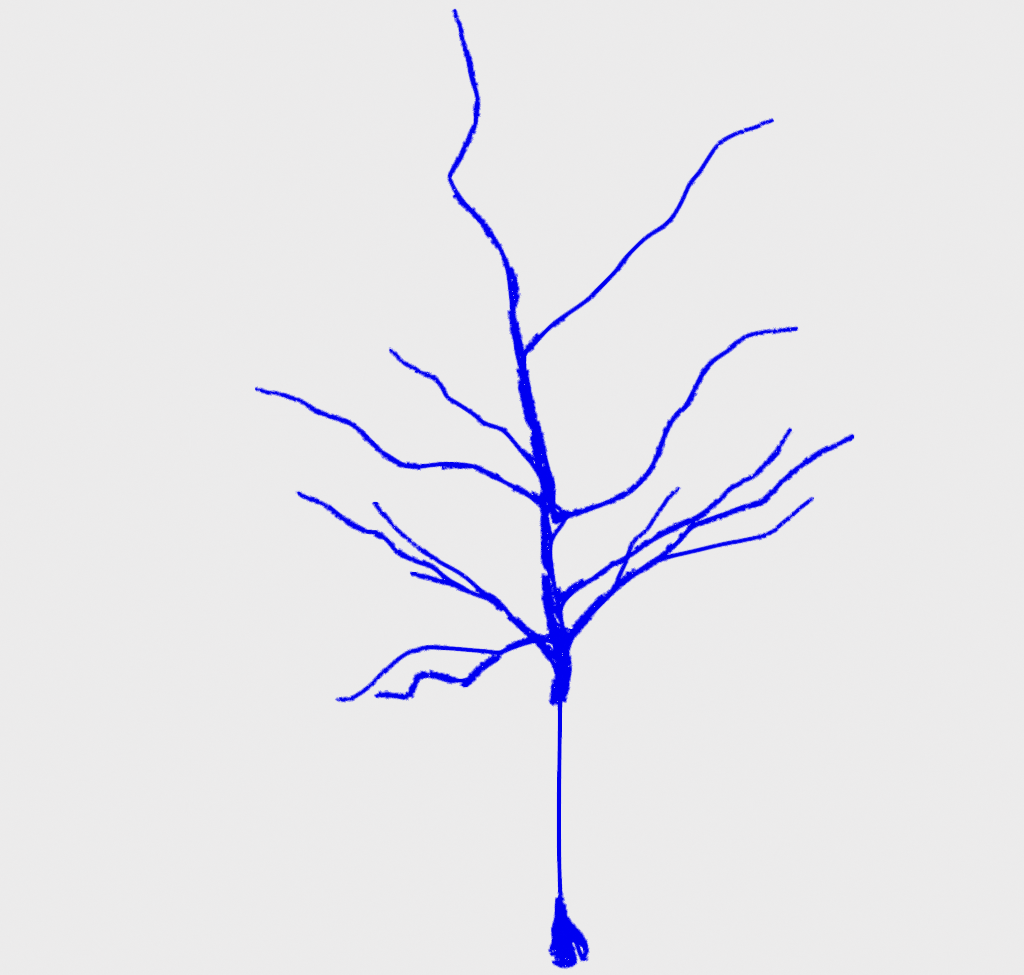}  \\
  \includegraphics[width=0.24\textwidth]{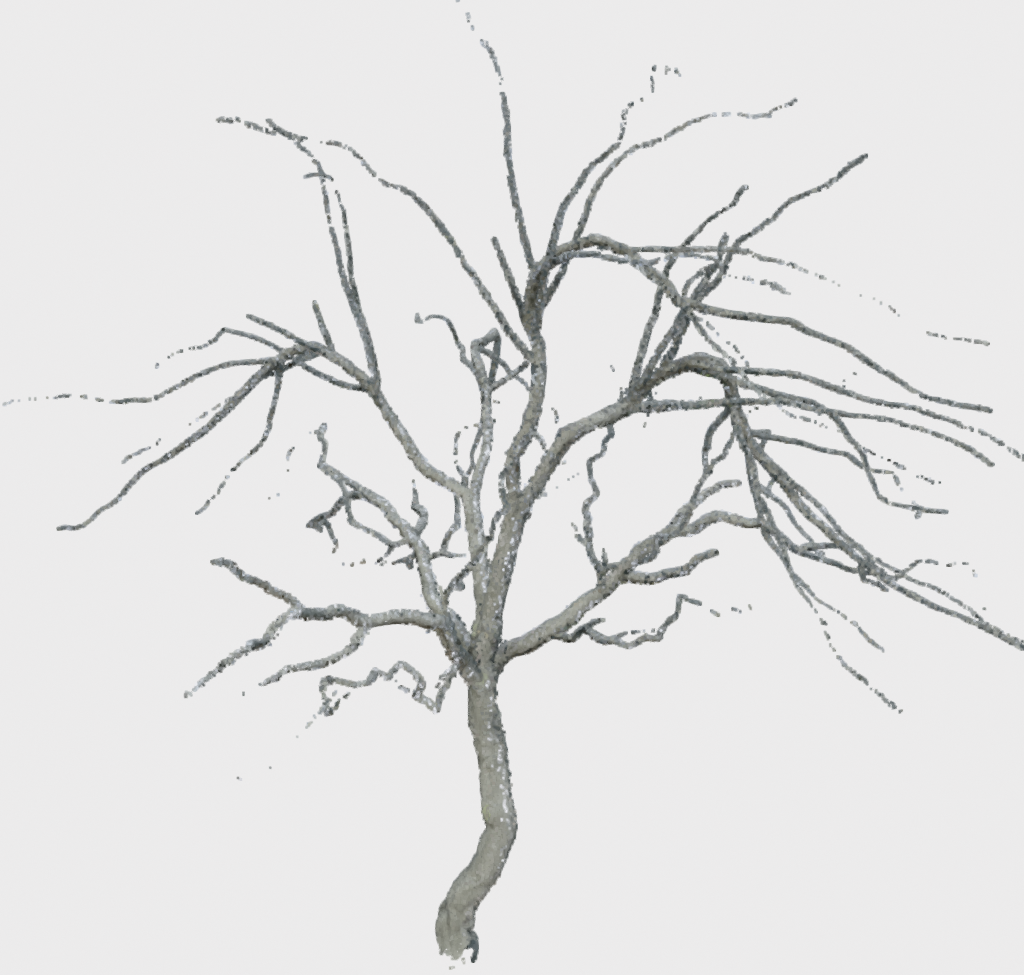} &  
  \includegraphics[width=0.24\textwidth]{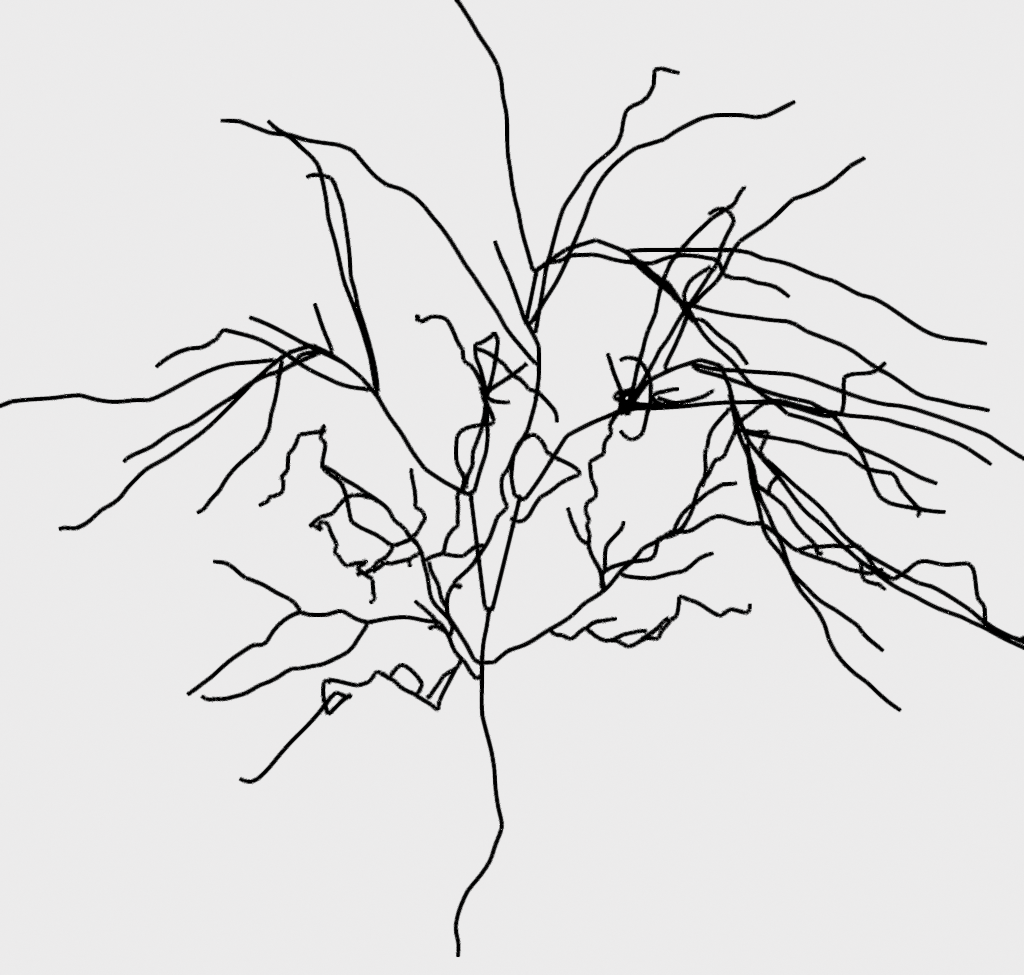}  &
  \includegraphics[width=0.24\textwidth]{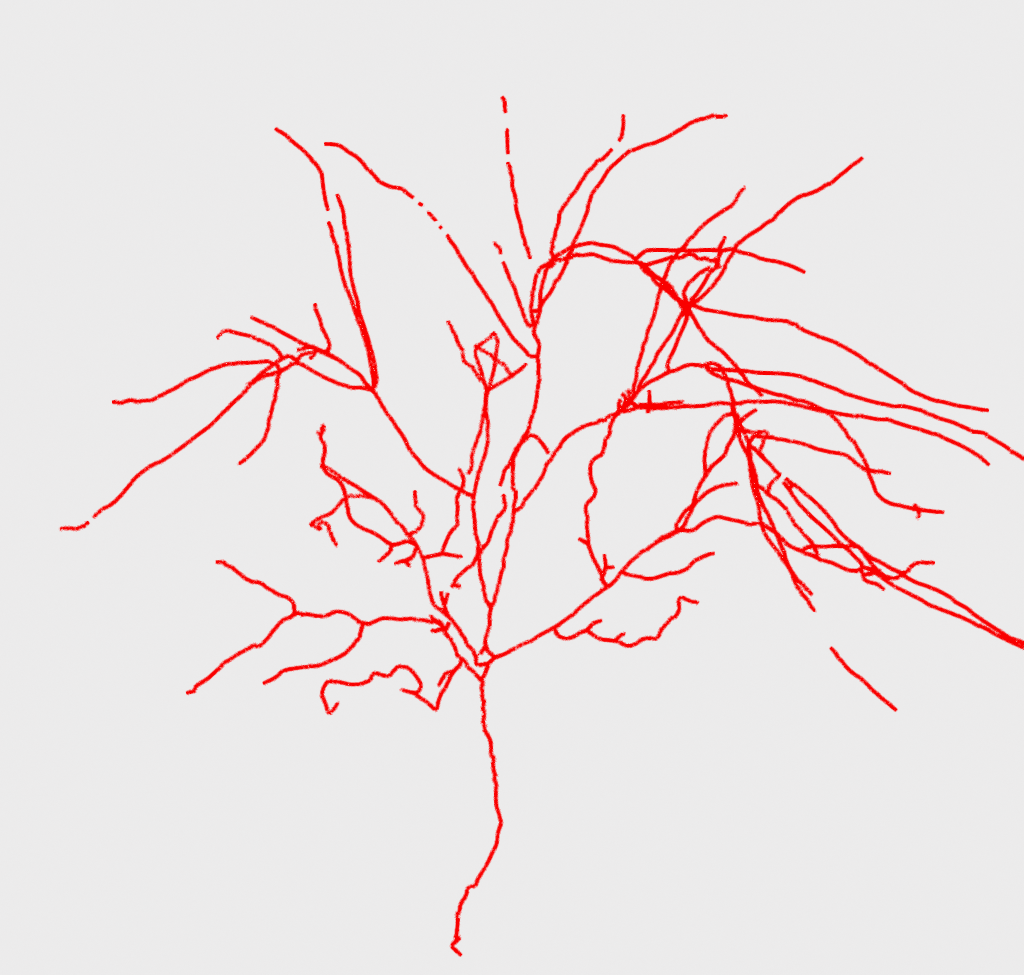}  &
  \includegraphics[width=0.24\textwidth]{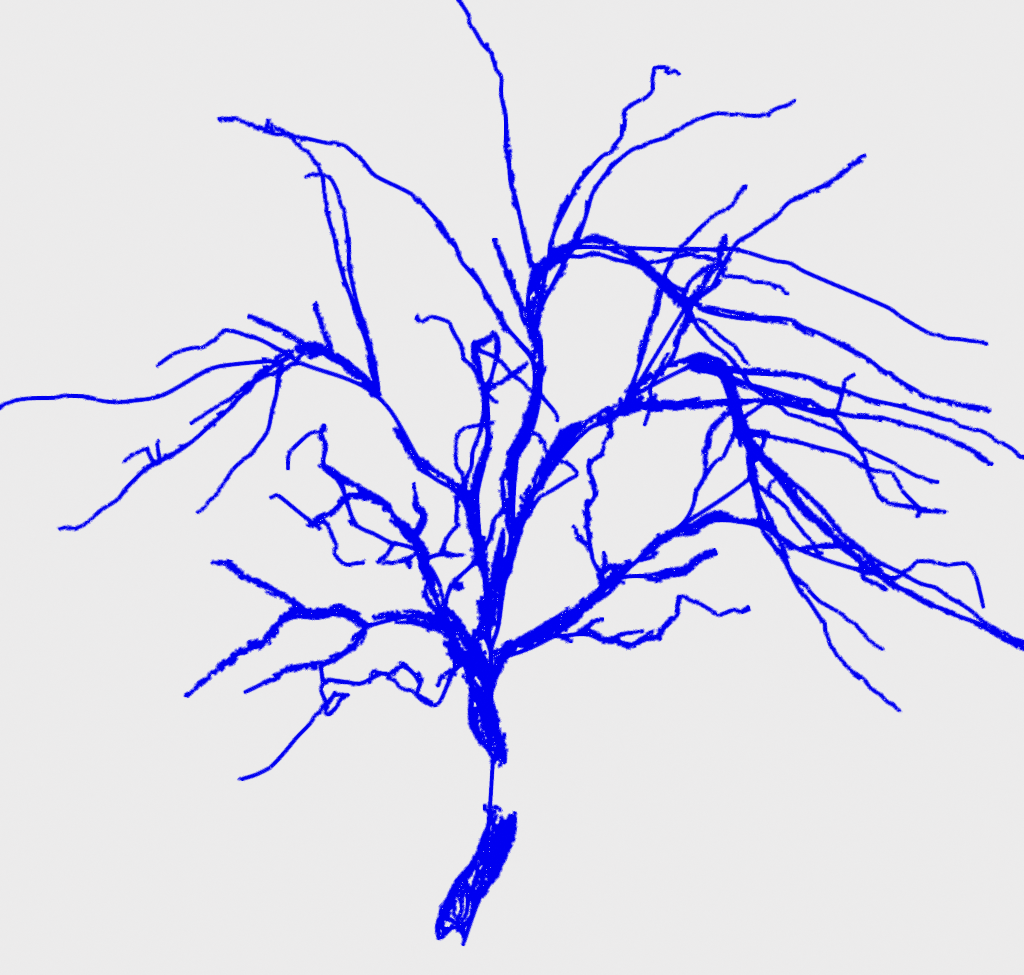}  \\
  \includegraphics[width=0.24\textwidth]{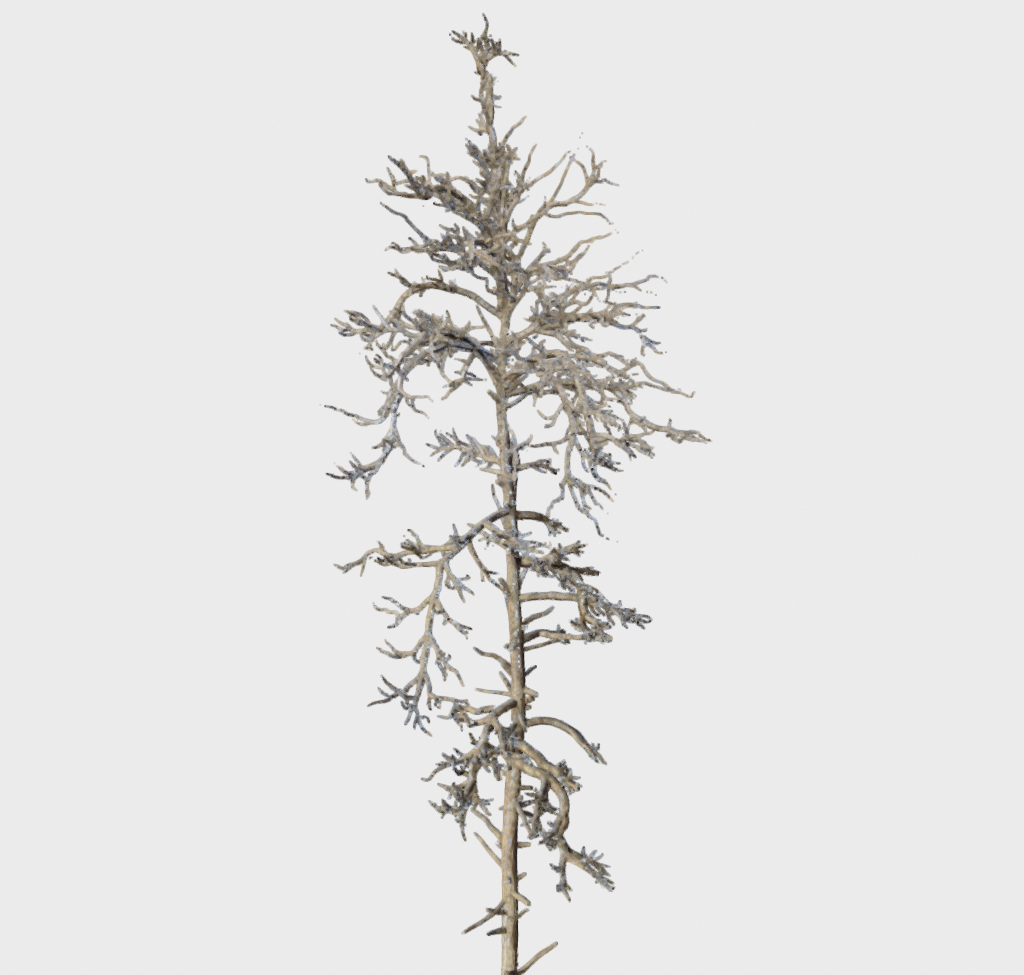} &  
  \includegraphics[width=0.24\textwidth]{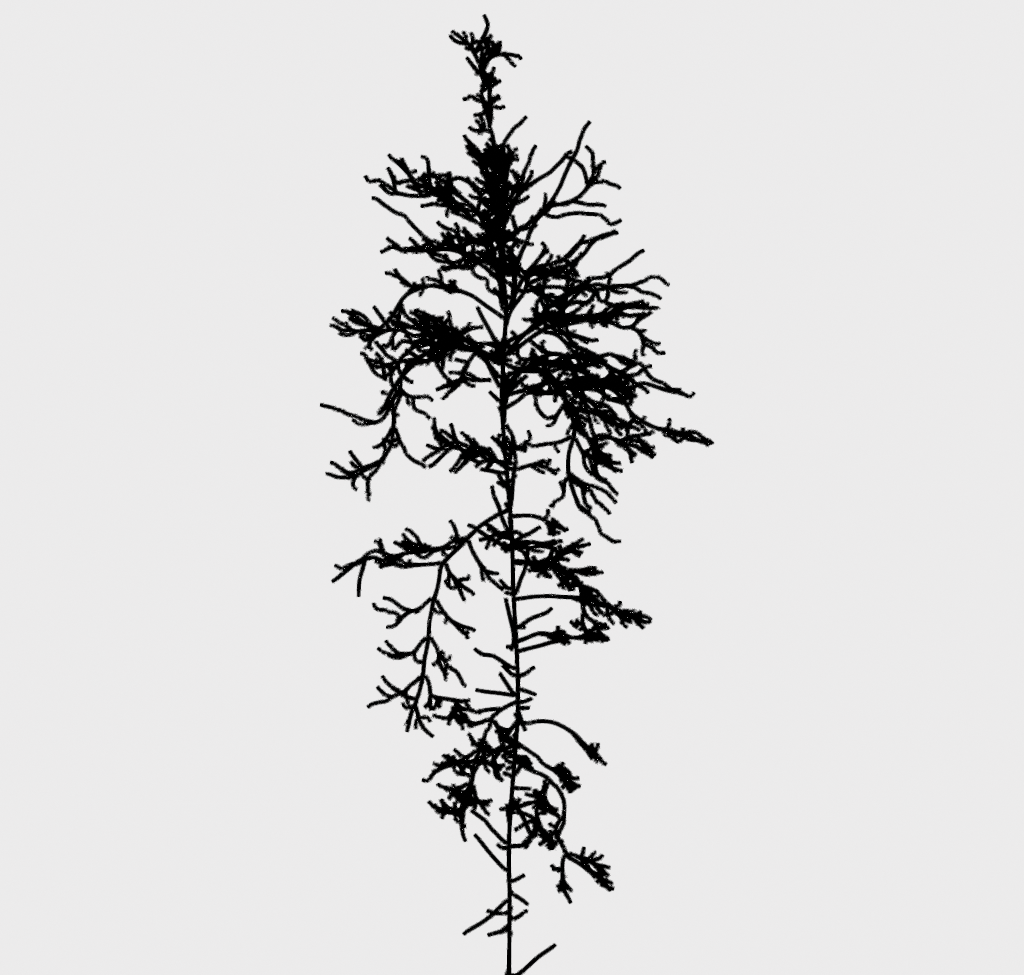}  &
  \includegraphics[width=0.24\textwidth]{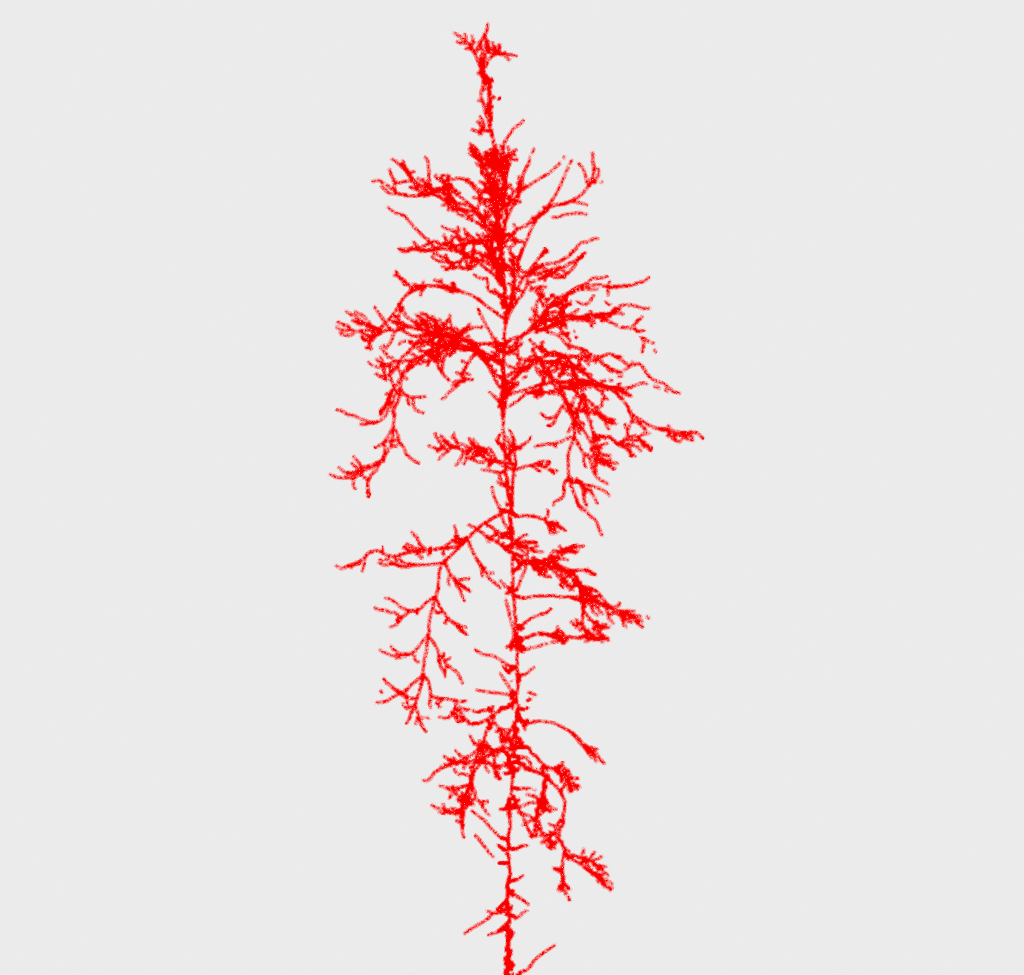}  &
  \includegraphics[width=0.24\textwidth]{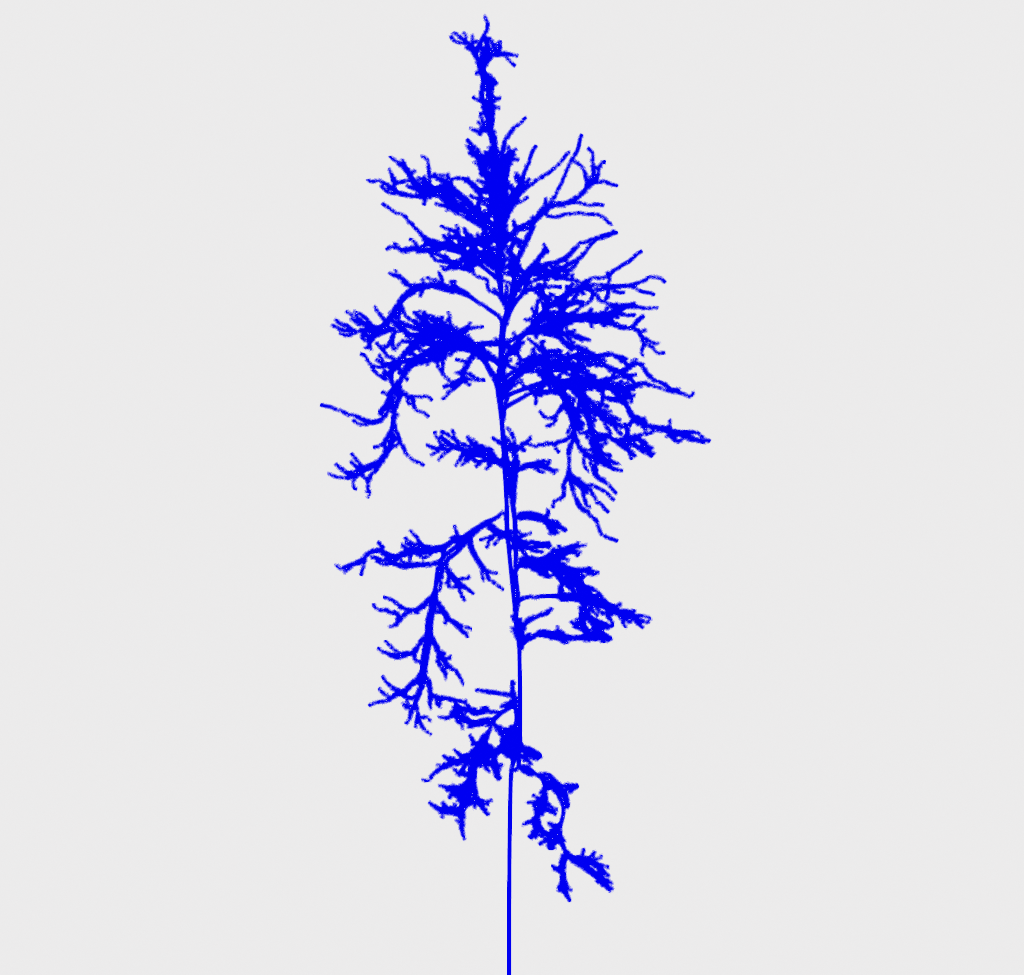}  \\
  \includegraphics[width=0.24\textwidth]{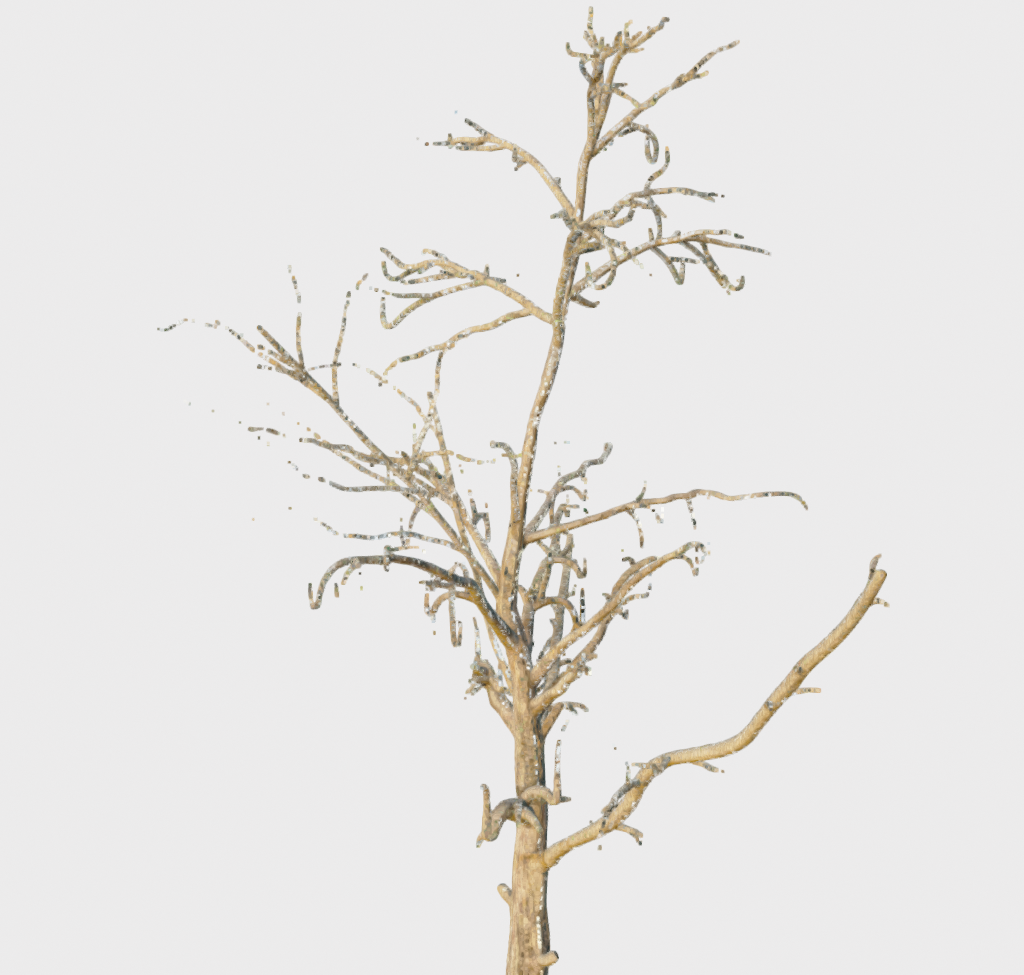} &  
  \includegraphics[width=0.24\textwidth]{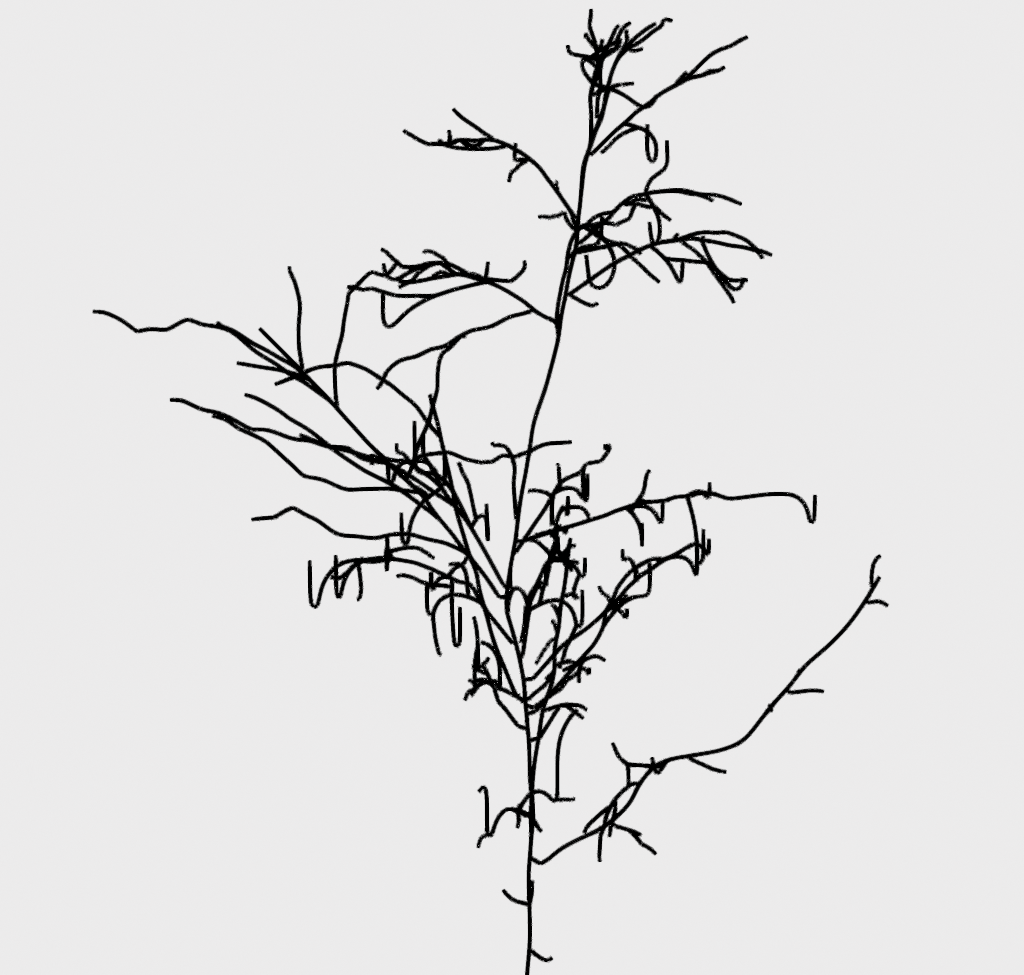}  &
  \includegraphics[width=0.24\textwidth]{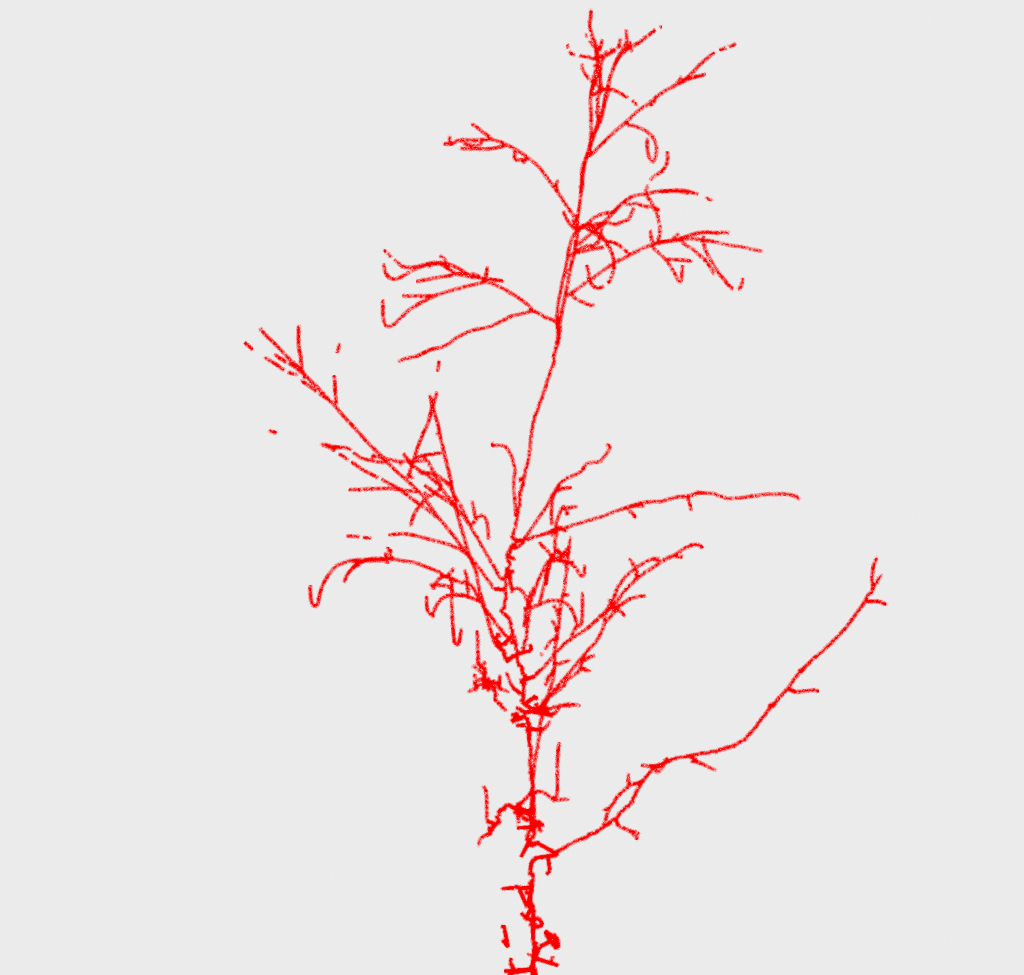}  &
  \includegraphics[width=0.24\textwidth]{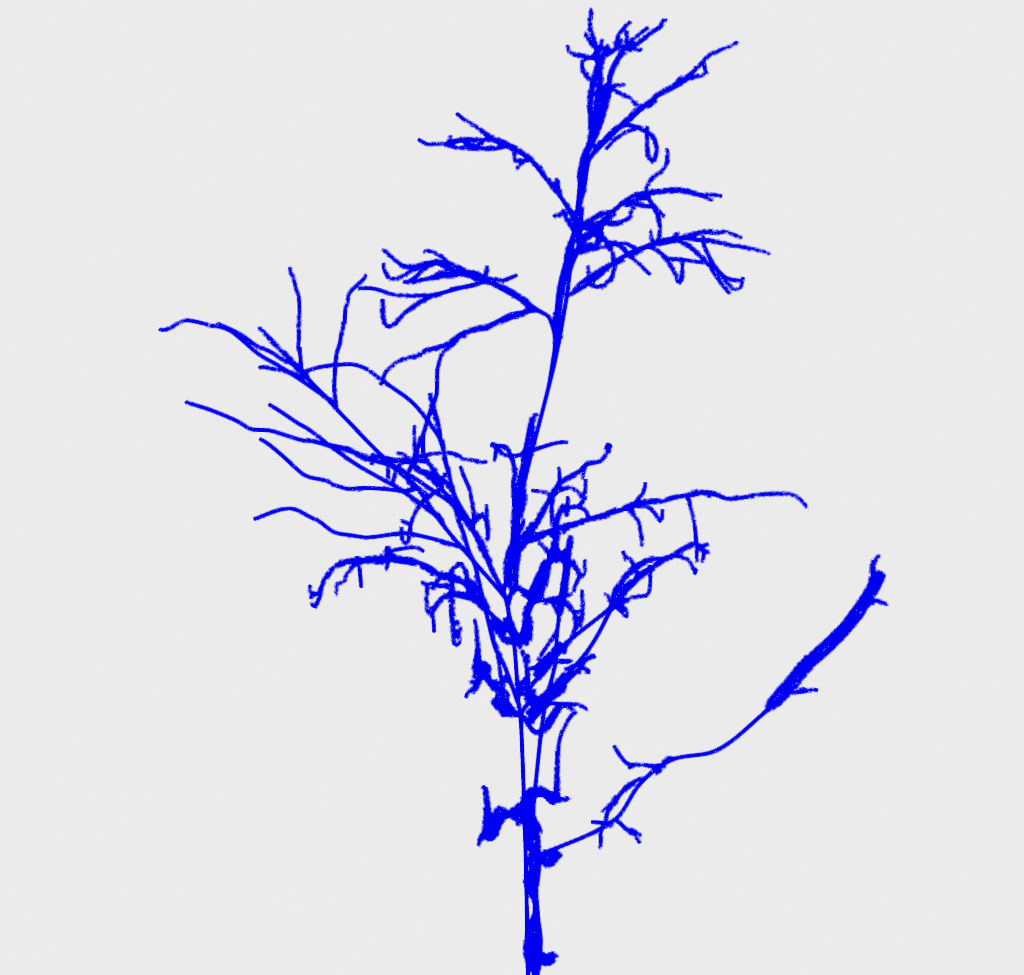}  \\
\end{tabular}
\caption{Several results of skeletonization of synthetic point clouds. From left to right: synthetic point cloud, ground truth skeleton, Smart-Tree skeleton, and Adtree skeleton. From top to bottom (species): cherry, eucalyptus, walnut, apple, pine, ginkgo}
\label{fig:qualitiative}
\vspace{-1.2cm}
\end{figure}

In Figure \ref{fig:qualitiative}, we show qualitative results for each species in our synthetic dataset. We can see that Smart-Tree produces an accurate skeleton representing the tree topology well. Adtree produces additional branches that would require post-processing to remove. 

AdTree often fails to capture the correct topology of the tree. This is due to Adtree using Delaunay triangulation to initialise the neighbourhood graph. This can lead to branches that have no association being connected. 

Smart-Tree, however, does not generate a fully connected skeleton - but rather one with multiple sub-graphs. The biggest sub-graph can still capture the majority of the major branching structure, although to provide a full topology of the tree with the finer branches, additional work is required to connect the sub-graphs by inferring the branching structure in occluded regions. 

\begin{figure}[H]
\centering
\begin{tabular}{ccc}
\includegraphics[width=0.29\textwidth, trim={10cm 0cm 12cm 0cm},clip]{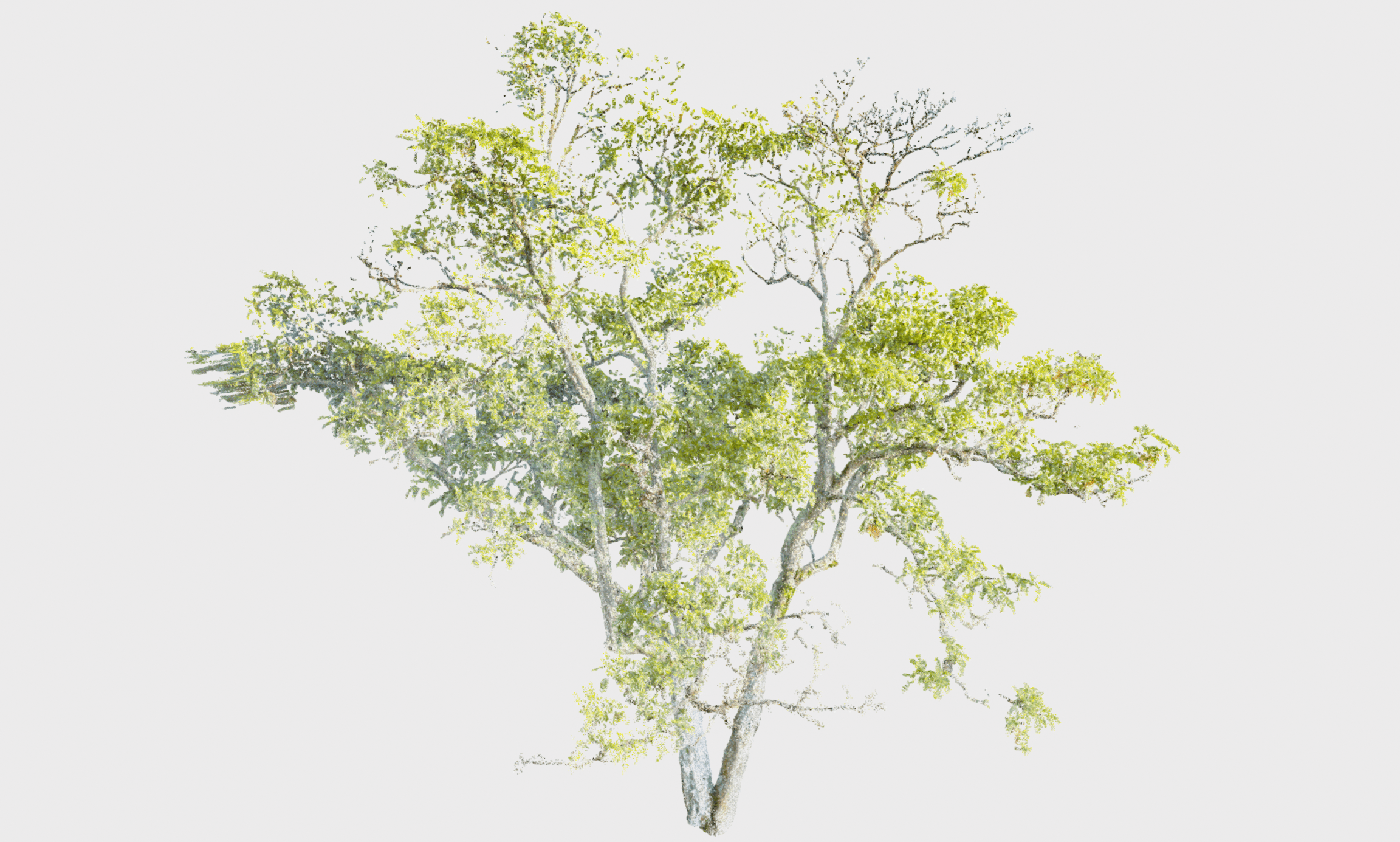}&
\includegraphics[width=0.29\textwidth, trim={10cm 0cm 12cm 0cm},clip]{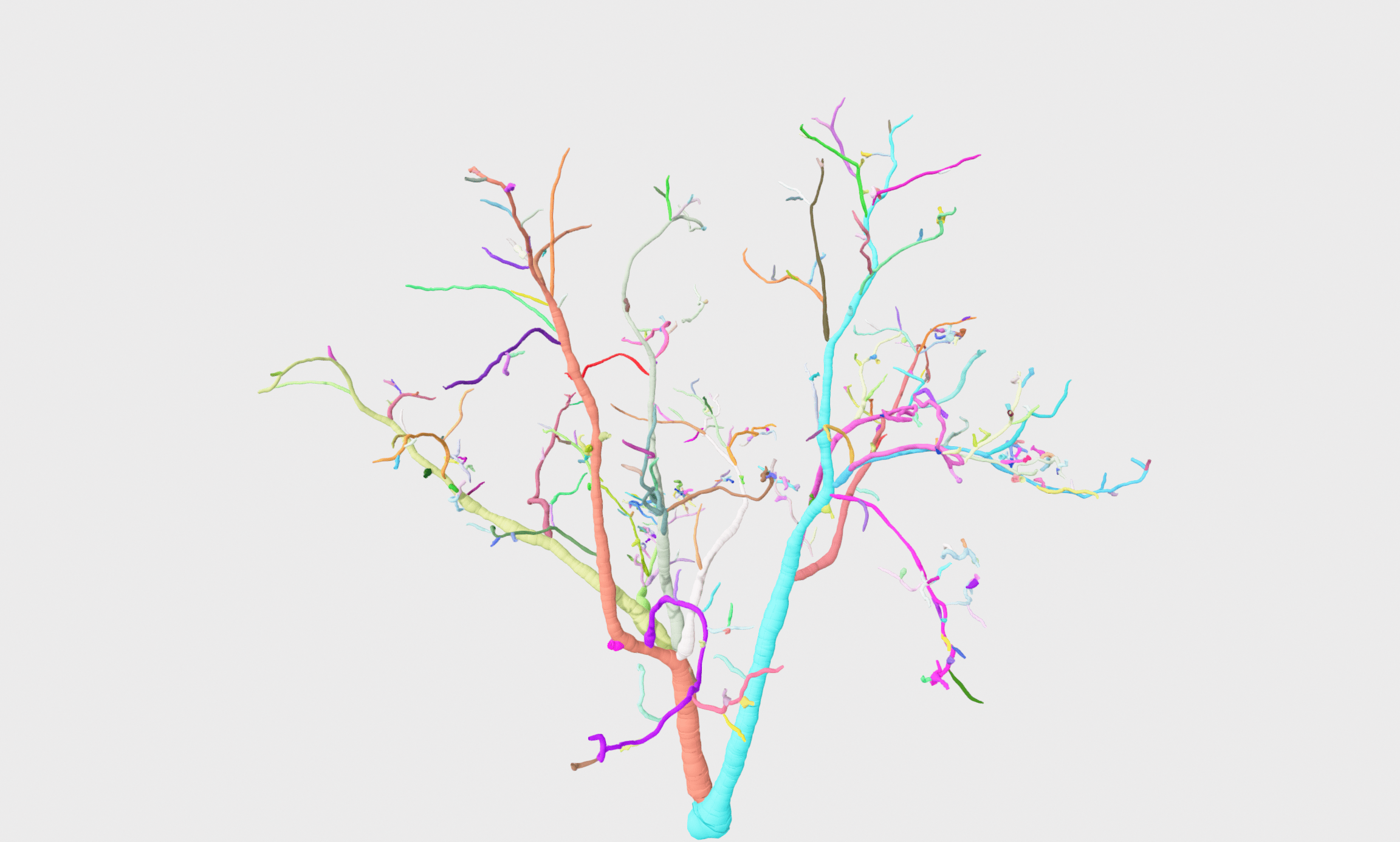}& 
\includegraphics[width=0.29\textwidth, trim={10cm 0cm 12cm 0cm},clip]{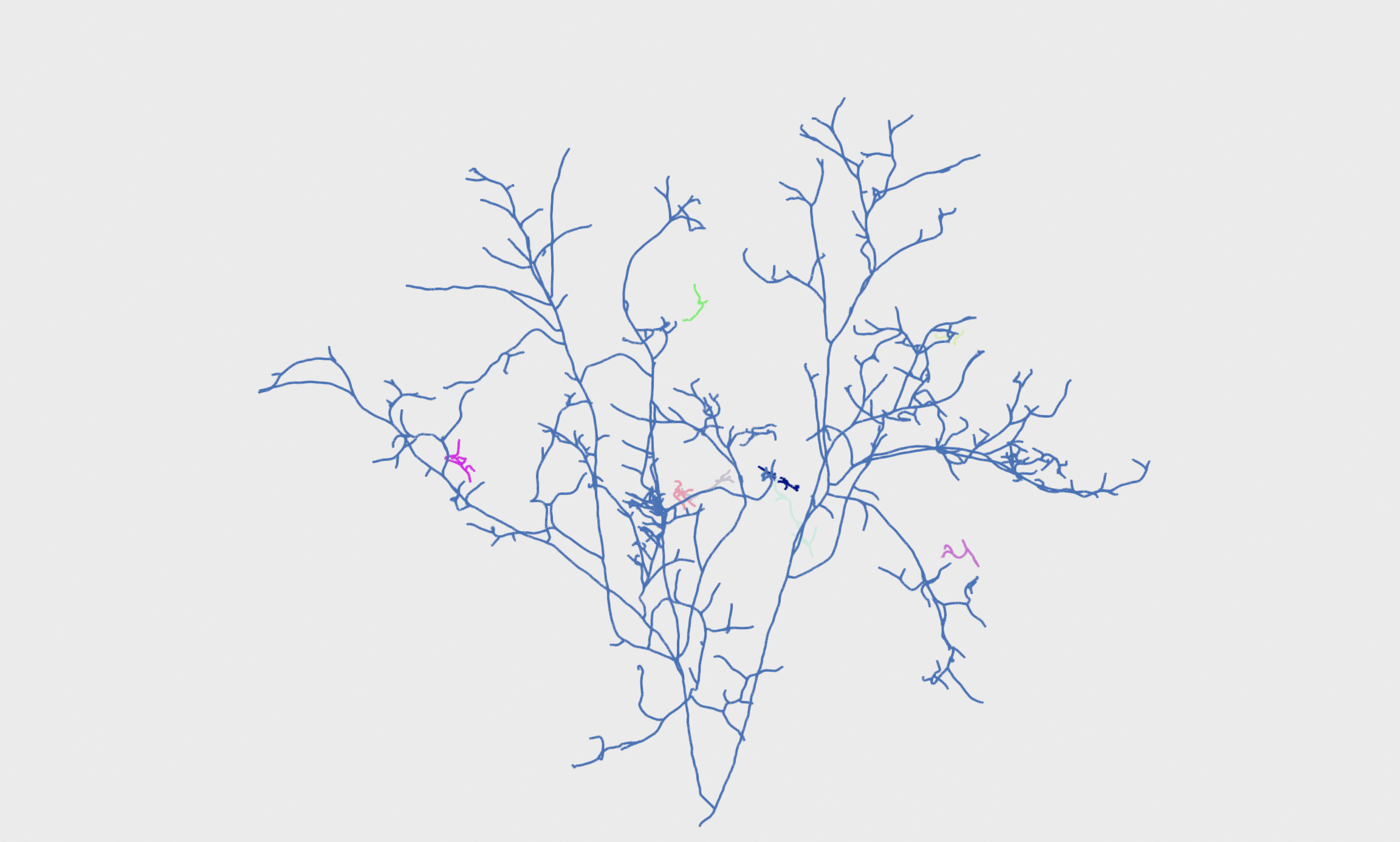}\\ 
\end{tabular}
\caption{Left to Right: Input point cloud (a), Branch meshes (b), Skeleton Sub-graphs (c).}
\label{fig:botanic-real}
\vspace{-10mm}%Put here to reduce too much white space after your table 
\end{figure}

%\subsection{Real Data}
To demonstrate our method's ability to work on real-world data. We test our method on a tree from the Christchurch Botanic Gardens, New Zealand. As this tree has foliage, we train our network to segment away the foliage points and then run the skeletonization algorithm on the remaining points. In Figure \ref{fig:botanic-real}c, we can see that Smart-Tree can accurately reconstruct the skeleton. 

% To demonstrate our method's ability to work on real-life data. We test our method on a Kowhai tree point cloud that was captured using a Phantom 4 RTK Drone (Figure \ref{fig:kowhai-real}).

% As the Kowhai tree contains foliage points, we train our network to segment the leaves from the wood as well as estimate the direction and radius. In Figure \ref{fig:kowhai-real}c, we can see that Smart-Tree can accurately reconstruct the skeleton. 

% \begin{figure}[H]
% \centering
% \begin{tabular}{ccc}
%   \includegraphics[width=32mm]{images/kowhai-pcd.png} &   \includegraphics[width=32mm]{images/kowhai-surface.png} & \includegraphics[width=32mm]{images/kowhai-skel.png} \\
% \end{tabular}
% \caption{Left to Right: Input point cloud (a), Allocated branch surface points (b), Skeleton output (c).}
% \label{fig:kowhai-real}
% %\vspace{-10mm}%Put here to reduce too much white space after your table 
% \end{figure}

% Furthermore, we qualitatively compare our method with Adtree, using a real-life apple tree dataset supplied by Straub et al. \cite{straub2022approach}.

\section{Conclusion and Future Work} \label{conclusion_section}
We proposed Smart-Tree, a supervised method for generating skeletons from tree point clouds. A major area for improvement in the literature on tree point-cloud skeletonization is quantitative evaluation, which we contribute towards with a synthetic tree point cloud dataset with ground-truth and error metrics.

We demonstrated that using a sparse convolutional neural network can help improve the robustness of tree point cloud skeletonization. One novelty of our work is that a neighbourhood graph can be created based on the radius at each region, improving the accuracy of our skeleton. 

We used a precision and recall-based metric to compare Smart-Tree with the state-of-the-art AdTree. Smart-Tree is generally much more precise than AdTree, but it currently does not handle point clouds containing gaps (due to occlusions and reconstruction errors). AdTree has problems with over-completeness on this dataset, with many duplicate branches. 

In the future, we would like to work towards robustness to gaps in the point cloud by filling gaps in the medial-axis estimation phase. We plan to train our method on a wider range of synthetic and real trees; to do this, we will expand our dataset to include more variety, trees with foliage, and human annotation on real trees. We are also working towards error metrics which better capture topology errors.

%\section*{Code Availability} \label{code_availability}

\section*{Acknowledgment}
This work was funded by the New Zealand Ministry of Business, Innovation and Employment under contract C09X1923 (Catalyst: Strategic Fund).
\newline
This research/project is supported by the National Research Foundation, Singapore under its Industry Alignment Fund – Pre-positioning (IAF-PP) Funding Initiative. Any opinions, findings and conclusions or recommendations expressed in this material are those of the author(s) and do not reflect the views of the National Research Foundation, Singapore.

\bibliographystyle{splncs04}

\bibliography{ref}

\end{document}